\documentclass{article}

\PassOptionsToPackage{numbers, compress}{natbib}

\usepackage[preprint]{nips_2018}

\usepackage[utf8]{inputenc} 
\usepackage[T1]{fontenc}    
\usepackage{hyperref}       
\usepackage{url}            
\usepackage{booktabs}       
\usepackage{amsfonts}       
\usepackage{nicefrac}       
\usepackage{microtype}      
\usepackage{amsmath}
\usepackage{enumerate}
\usepackage{graphicx}
\usepackage[titlenumbered,ruled]{algorithm2e}
\usepackage{comment}
\usepackage[symbol]{footmisc}
\RequirePackage{authblk}

\newcommand{\networkname}{Hierarchical Relation Network}
\newcommand{\networkshort}{HRN}
\newcommand{\mpos}{X}
\newcommand{\lmpos}{x}
\newcommand{\R}{\mathbb{R}}

\title{Flexible Neural Representation for Physics Prediction}
\author[ ]{Damian Mrowca$^{1,}\footnote[1]{}$}
\author[ ]{Chengxu Zhuang$^{2,}\footnote[1]{}$}
\author[ ]{Elias Wang$^{3,}\footnote[1]{}$}
\author[ ]{Nick Haber$^{2,4,5}$}
\author[ ]{Li Fei-Fei$^1$}
\author[ ]{\\Joshua B. Tenenbaum$^{7, 8}$}
\author[ ]{Daniel L. K. Yamins$^{1, 2, 6}$}

\affil[ ]{Department of Computer Science$^1$, Psychology$^2$, Electrical Engineering$^3$, Pediatrics$^4$ and Biomedical Data Science$^5$, and Wu Tsai Neurosciences Institute$^6$, Stanford, CA 94305}

\affil[ ]{Department of Brain and Cognitive Sciences$^7$, and Computer Science and Artificial Intelligence Laboratory$^8$, MIT, Cambridge, MA 02139 \vspace{0.15cm}}

\affil[ ]{\texttt{\{\href{mailto:mrowca@stanford.edu}{mrowca}, \href{mailto:chengxuz@stanford.edu}{chengxuz}, \href{mailto:eliwang@stanford.edu}{eliwang}\}@stanford.edu}}

\begin{document}
\maketitle

\footnotetext[1]{Equal contribution}

\begin{abstract}
Humans have a remarkable capacity to understand the physical dynamics of objects in their environment, flexibly capturing complex structures and interactions at multiple levels of detail.  
Inspired by this ability, we propose a hierarchical particle-based object representation that covers a wide variety of types of three-dimensional objects, including both arbitrary rigid geometrical shapes and deformable materials.  
We then describe the Hierarchical Relation Network (HRN), an end-to-end differentiable neural network based on hierarchical graph convolution, that learns to predict physical dynamics in this representation. 
Compared to other neural network baselines, the HRN accurately handles complex collisions and nonrigid deformations, generating plausible dynamics predictions at long time scales in novel settings, and scaling to large scene configurations.
These results demonstrate an architecture with the potential to form the basis of next-generation physics predictors for use in computer vision, robotics, and quantitative cognitive science. 
\end{abstract}

\section{Introduction}
Humans efficiently decompose their environment into objects, and reason effectively about the dynamic interactions between these objects~\citep{spelke1992origins, tenenbaum2011grow}. Although human intuitive physics may be quantitatively inaccurate under some circumstances~\citep{mccloskey1980curvilinear}, humans make qualitatively plausible guesses about dynamic trajectories of their environments over long time horizons \citep{smith2013sources}.  Moreover, they either are born knowing, or quickly learn about, concepts such as object permanence, occlusion, and deformability, which guide their perception and reasoning \citep{spelke1990principles}.

An artificial system that could mimic such abilities would be of great use for applications in computer vision, robotics, reinforcement learning, and many other areas. While traditional physics engines constructed for computer graphics have made great strides, such routines are often hard-wired and thus challenging to integrate as components of larger learnable systems. Creating end-to-end differentiable neural networks for physics prediction is thus an appealing idea. Recently, \citet{chang2016compositional} and \citet{battaglia2016interaction} have illustrated the use of neural networks to predict physical object interactions in (mostly) 2D scenarios by proposing object-centric and relation-centric representations. Common to these works is the treatment of scenes as graphs, with nodes representing object point masses and edges describing the pairwise relations between objects (e.g. gravitational, spring-like, or repulsing relationships). Object relations and physical states are used to compute the pairwise effects between objects. After combining effects on an object, the future physical state of the environment is predicted on a per-object basis. This approach is very promising in its ability to explicitly handle object interactions.  However, a number of challenges have remained in generalizing this approach to real-world physical dynamics, including representing arbitrary geometric shapes with sufficient resolution to capture complex collisions, working with objects at different scales simultaneously, and handling non-rigid objects of nontrivial complexity. 

\begin{figure}[!ht]
  \centering
  \label{fig:setup}
  \vspace{-0.2cm}
  \includegraphics[width=\textwidth]{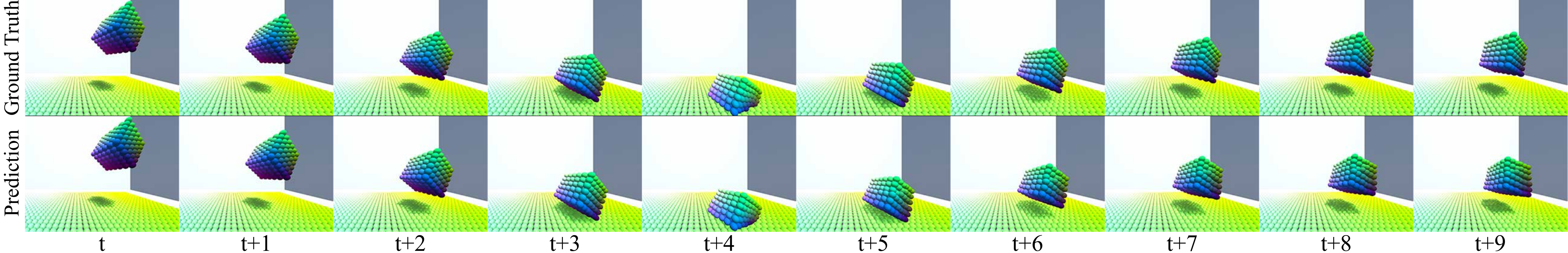}
  \caption{\textbf{Predicting physical dynamics.} Given past observations the task is to predict the future physical state of a system. In this example, a cube deforms as it collides with the ground. The top row shows the ground truth and the bottom row the prediction of our physics prediction network.}
  \vspace{-0.2cm}
\end{figure}

Several of these challenges are illustrated in the fast-moving deformable cube sequence depicted in Figure \ref{fig:setup}. Humans can flexibly vary the level of detail at which they perceive such objects in motion: The cube may naturally be conceived as an undifferentiated point mass as it moves along its initial kinematic trajectory. But as it collides with and bounces up from the floor, the cube's complex rectilinear substructure and nonrigid material properties become important for understanding what happens and predicting future interactions.  The ease with which the human mind handles such complex scenarios is an important explicandum of cognitive science, and also a key challenge for artificial intelligence.  Motivated by both of these goals, our aim here is to develop a new class of neural network architectures with this human-like ability to reason flexibly about the physical world. 

To this end, it would be natural to extend the interaction network framework by representing each object as a (potentially large) set of connected particles. In such a representation, individual constituent particles could move independently, allowing the object to deform while being constrained by pairwise relations preventing the object from falling apart. However, this type of particle-based representation introduces a number of challenges of its own.  Conceptually, it is not immediately clear how to efficiently propagate effects across such an object.  Moreover, representing every object with hundreds or thousands of particles would result in an exploding number of pairwise relations, which is both computationally infeasible and cognitively unnatural.

As a solution to these issues, we propose a novel cognitively-inspired hierarchical graph-based object representation that captures a wide variety of complex rigid and deformable bodies (Section 3), and an efficient hierarchical graph-convolutional neural network that learns physics prediction within this representation (Section 4). Evaluating on complex 3D scenarios, we show substantial improvements relative to strong baselines both in quantitative prediction accuracy and qualitative measures of prediction plausibility, and evidence for generalization to complex unseen scenarios (Section 5).

\section{Related Work}
An efficient and flexible predictor of physical dynamics has been an outstanding question in neural network design. 
In computer vision, modeling moving objects in images or videos for action recognition, future prediction, and object tracking is of great interest. 
Similarly in robotics, action-conditioned future prediction from images is crucial for navigation or object interactions. 
However, future predictors operating directly on 2D image representations often fail to generate sharp object boundaries and struggle with occlusions and remembering objects when they are no longer visually observable~\citep{agrawal2016learning, fragkiadaki2015learning, finn2016unsupervised, lerer2016learning, li2016fall, mottaghi2016newtonian, mottaghi2016happens, haber2018learning}. 
Representations using 3D convolution or point clouds are better at maintaining object shape~\citep{tran2015learning, tran2016deep, byravan2017se3, qi2017pointnet, qi2017pointnet++}, but do not entirely capture object permanence, and can be computationally inefficient. More similar to our approach are inverse graphics methods that extract a lower dimensional physical representation from images that is used to predict physics \citep{kulkarni2014inverse, kulkarni2015deep, whitney2016understanding, watters2017visual, wu2015galileo, wu2016physics, brand1997physics, wang20183d}. Our work draws inspiration from and extends that of \citet{chang2016compositional} and \citet{battaglia2016interaction}, which in turn use ideas from graph-based neural networks \citep{scarselli2009graph, sutskever2009using, bruna2013spectral, li2015gated, henaff2015deep, duvenaud2015convolutional, defferrard2016convolutional, kipf2016semi, bronstein2017geometric, schlichtkrull2017modeling}. 
Most of the existing work, however, does not naturally handle complex scene scenarios with objects of widely varying scales or deformable objects with complex materials.

Physics simulation has also long been studied in computer graphics, most commonly for rigid-body collisions \citep{baraff2001physically, coumans2010bullet}. 
Particles or point masses have also been used to represent more complex physical objects, with the neural network-based NeuroAnimator being one of the earliest examples to use a hierarchical particle representation for objects to advance the movement of physical objects \citep{grzeszczuk1998neuroanimator}. 
Our particle-based object representation also draws inspiration from recent work on (non-neural-network) physics simulation, in particular the NVIDIA FleX engine~\citep{macklin2014unified, bender2015position}. 
However, unlike this work, our solution is an end-to-end differentiable neural network that can learn from data.

Recent research in computational cognitive science has posited that humans run physics simulations in their mind \citep{battaglia2013simulation, bates2015humans, hamrick2011internal, ullman2014learning, hegarty2004mechanical}. 
It seems plausible that such simulations happen at just the right level of detail which can be flexibly adapted as needed, similar to our proposed representation. 
Both the ability to imagine object motion as well as to flexibly decompose an environment into objects and parts form an important prior that humans rely on for further learning about new tasks, when generalizing their skills to new environments or flexibly adapting to changes in inputs and goals \citep{lake2017building}.

\section{Hierarchical Particle Graph Representation}
A key factor for predicting the future physical state of a system is the underlying representation used.
A simplifying, but restrictive, often made assumption is that all objects are rigid. 
A rigid body can be represented with a single point mass and unambiguously situated in space by specifying its position and orientation, together with a separate data structure describing the object's shape and extent.
Examples are 3D polygon meshes or various forms of 2D or 3D masks extracted from perceptual data \citep{byravan2017se3, finn2016unsupervised}.
The rigid body assumption describes only a fraction of the real world, excluding, for example, soft bodies, cloths, fluids, and gases, and precludes objects breaking and combining.
However, objects are divisible and made up of a potentially large numbers of smaller sub-parts.

Given a scene with a set of objects $O$, the core idea is to represent each object $o \in O$ with a set of particles $P_{o} \equiv \{p_i| i \in o \}$. Each particle's state at time $t$ is described by a vector in $\R^7$ consisting of its position $x \in \R^{3}$, velocity $\delta \in \R^{3}$, and mass $m \in \R^+$. We refer to $p_i$ and this vector interchangeably.

Particles are spaced out across an object to fully describe its volume. In theory, particles can be arbitrarily placed within an object. Thus, less complex parts can be described with fewer particles (e.g. 8 particles fully define a cube). More complicated parts (e.g. a long rod) can be represented with more particles.
We define $P$ as the set $\{p_i | 1 \leq i \leq N_P\}$ of all $N_P$ particles in the observed scene.

\begin{figure}[!ht]
  \centering
  \vspace{-0.2cm}
  \includegraphics[width=\textwidth]{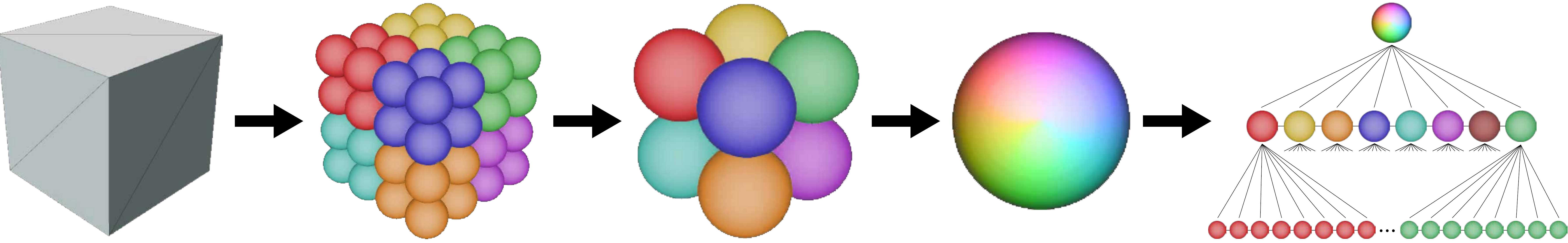}
  \caption{\textbf{Hierarchical graph-based object representation}. An object is decomposed into particles. Particles (of the same color) are grouped into a hierarchy representing multiple object scales. Pairwise relations constrain particles in the same group and to ancestors and descendants.}
  \label{fig:particle_representation}
  \vspace{-0.2cm}
\end{figure}

To fully physically describe a scene containing multiple objects with particles, we also need to define how the particles relate to each other.
Similar to \citet{battaglia2016interaction}, we represent relations between particles $p_i$ and $p_j$ with $K$-dimensional pairwise relationships $R = \{r_{ij} \in \R^K\}$.
Each relationship $r_{ij}$ within an object encodes material properties. For example, for a soft body $r_{ij} \in \R$ represents the local material stiffness, which need not be uniform within an object. Arbitrarily-shaped objects with potentially nonuniform materials can be represented in this way. Note that the physical interpretation of $r_{ij}$ is learned from data rather than hard-coded through equations. 
Overall, we represent the scene by a node-labeled graph $G = \langle P, R \rangle$ where the particles form the nodes $P$ and the relations define the (directed) edges $R$. Except for the case of collisions, different objects are disconnected components within $G$.

The graph $G$ is used to propagate effects through the scene. It is infeasible to use a fully connected graph for propagation as pairwise-relationship computations grow with $\mathcal O(N_P^2)$. To achieve $\mathcal{O}(N_Plog(N_P))$ complexity, we construct a {\em hierarchical scene (di)graph} $G_H$ from $G$ in which the nodes of each connected component are organized into a tree structure: 
First, we initialize the leaf nodes $L$ of $G_H$ as the original particle set $P$.
Then, we extend $G_H$ by a root node for each connected component (object) in $G$. The root node states are defined as the aggregates of their leaf node states. The root nodes are connected to their leaves with directed edges and vice versa. 

At this point, $G_H$ consists of the leaf particles $L$ representing the finest scene resolution and one root node for each connected component describing the scene at the object level. To obtain intermediate levels of detail, we then cluster the leaves $L$ in each connected component into smaller subcomponents using a modified k-means algorithm. We add one node for each new subcomponent and connect its leaves to the newly added node and vice versa. This newly added node is then labeled as the direct ancestors for its leaves and its leaves are siblings to each other. We then connect the added intermediate nodes with each other if and only if their respective subcomponent leaves are connected. Lastly, we add directed edges from the root node of each connected component to the new intermediate nodes in that component, and remove edges between leaves not in the same cluster.
The process then recurses within each new subcomponent. See Algorithm \ref{fig:grouping_algorithm} in the supplementary for details.

We denote the sibling(s) of a particle $p$ by $\textrm{sib}(p)$, its ancestor(s) by $\textrm{anc}(p)$, its parent by $\textrm{par}(p)$, and its descendant(s) by $\textrm{des}(p)$. We define $\textrm{leaves}(p_a) = \{p_l \in L \ | \ p_a \in \textrm{anc}(p_l)\}$.
Note that in $G_H$, directed edges connect $p_i$ and $\textrm{sib}(p_i)$, leaves $p_l$ and $\textrm{anc}(p_l)$, and $p_i$ and $\textrm{des}(p_i)$; see Figure \ref{fig:graphconv}b. 

\section{Physics Prediction Model}
In this section we introduce our physics prediction model.
It is based on hierarchical graph convolution, an operation which propagates relevant physical effects through the graph hierarchy.

\subsection{Hierarchical Graph Convolutions For Effect Propagation}
\begin{figure}[!ht]
  \centering
  \vspace{-0.2cm}
  \includegraphics[width=\textwidth]{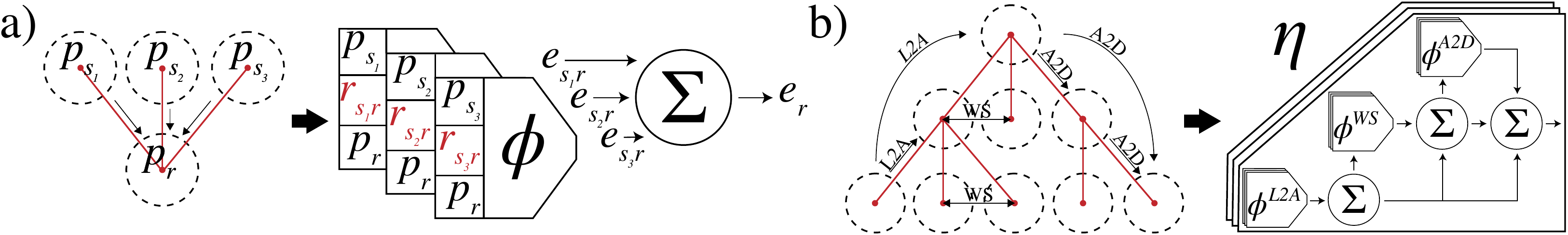}
  \caption{\textbf{Effect propagation through graph convolutions.} \textbf{a) Pairwise graph convolution $\phi$}. A receiver particle $p_r$ is constrained in its movement through graph relations $r_{sr}$ with sender particle(s) $p_s$. Given $p_s$, $p_r$ and $r_{sr}$, the effect $e_{sr}$ of $p_s$ on $p_r$ is computed using a fully connected neural network. The overall effect $e_r$ is the sum of all effects on $p_r$. \textbf{b) Hierarchical graph convolution $\eta$}. Effects in the hierarchy are propagated in three consecutive steps. (1) $\phi_{L2A}$. Leaf particles $L$ propagate effects to all of their ancestors $A$. (2) $\phi_{WS}$. Effects are exchanged between siblings $S$. (3) $\phi_{A2D}$. Effects are propagated from the ancestors $A$ to all of their descendants $D$.}
  \label{fig:graphconv}
  \vspace{-0.2cm}
\end{figure}

In order to predict the future physical state, we need to resolve the constraints that particles connected in the hierarchical graph impose on each other. 
We use graph convolutions to compute and propagate these effects.
Following \citet{battaglia2016interaction}, we implement a \emph{pairwise graph convolution} using two basic building blocks: (1) A pairwise processing unit $\phi$ that takes the sender particle state $p_s$, the receiver particle state $p_r$ and their relation $r_{sr}$ as input and outputs the effect $e_{sr} \in \R^E$ of $p_s$ on $p_r$, and (2) a commutative aggregation operation $\Sigma$ which collects and computes the overall effect $e_r \in \R^E$. In our case, this is a simple summation over all effects on $p_r$. Together these two building blocks form a convolution on graphs as shown in Figure~\ref{fig:graphconv}a.

Pairwise processing limits graph convolutions to only propagate effects between directly connected nodes. For a generic flat graph, we would have to repeatedly apply this operation until the information from all particles has propagated across the whole graph. This is infeasible in a scenario with many particles. Instead, we leverage direct connections between particles and their ancestors in our hierarchy to propagate all effects across the entire graph in \emph{one} model step. We introduce a \emph{hierarchical graph convolution}, a three stage mechanism for effect propagation as seen in Figure~\ref{fig:graphconv}b:

The first \emph{L2A (Leaves to Ancestors) stage} $\phi^{L2A}(p_l, p_a, r_{la} , e^0_l)$ 
predicts the effect $e^{L2A}_{la} \in \R^E$ of a leaf particle $p_l$ on an ancestor particle $p_a \in \textrm{anc}(p_l)$ 
given $p_l$, $p_a$, the material property information of $r_{la}$, and input effect $e^0_l$ on $p_l$.
The second \emph{WS (Within Siblings) stage} $\phi^{WS}(p_i, p_j, r_{ij}, e^{L2A}_i)$ predicts the effect $e^{WS}_{ij} \in \R^E$ of sibling particle $p_i$ on $p_j \in \textrm{sib}(p_i)$.
The third \emph{A2D (Ancestors to Descendants) stage} $\phi^{A2D}(p_a, p_d, r_{ad}, e^{L2A}_a + e^{WS}_a)$ predicts the effect $e^{A2D}_{ij} \in \R^E$ of an ancestor particle $p_a$ on a descendant particle $p_d \in \textrm{des}(p_a)$. 
The total propagated effect $e_i$ on particle $p_i$ is computed by summing the various effects on that particle, $e_i = e^{L2A}_i + e^{WS}_i + e^{A2D}_i$ where
\begin{equation*}
e^{L2A}_a = \sum_{p_l \in \textrm{leaves}(p_a)} \phi^{L2A}(p_l, p_a, r_{la}, e^0_l) \qquad \qquad
e^{WS}_j = \sum_{p_i \in \textrm{sib}(p_j)} \phi^{WS}(p_i, p_j, r_{ij}, e^{L2A}_i)
\end{equation*}
\begin{equation*}
e^{A2D}_d = \sum_{p_a \in \textrm{anc}(p_d)} \phi^{A2D}(p_a, p_d, r_{ad}, e^{L2A}_a + e^{WS}_a).
\end{equation*}

In practice, $\phi^{L2A}, \phi^{WS},$ and $\phi^{A2D}$ are realized as fully-connected networks with shared weights that receive an additional ternary input ($0$ for L2A, $1$ for WS, and $2$ for A2D) in form of a one-hot vector.

Since all particles within one object are connected to the root node, information can flow across the entire hierarchical graph in at most two propagation steps. We make use of this property in our model.

\subsection{The Hierarchical Relation Network Architecture}

\begin{figure}
  \centering
  \includegraphics[width=\textwidth]{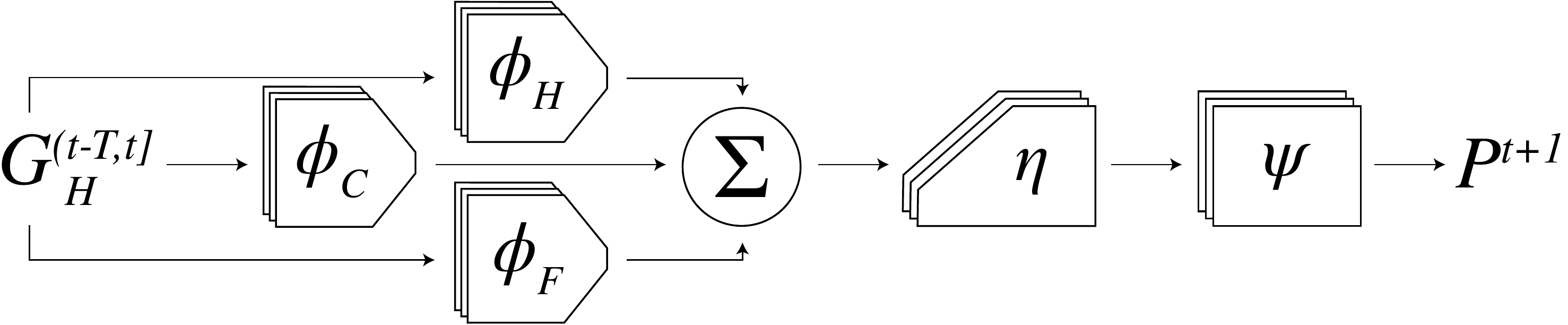}
  \caption{\textbf{Hierarchical Relation Network}. The model takes the past particle graphs $G_H^{(t-T, t]} = \langle P^{(t-T, t]}, R^{(t-T, t]} \rangle$ as input and outputs the next states $P^{t+1}$. The inputs to each graph convolutional effect module $\phi$ are the particle states and relations, the outputs the respective effects. $\phi_H$ processes past states, $\phi_C$ collisions, and $\phi_F$ external forces. The hierarchical graph convolutional module $\eta$ takes the sum of all effects, the pairwise particle states, and relations and propagates the effects through the graph. Finally, $\psi$ uses the propagated effects to compute the next particle states $P^{t+1}$.}
  \label{fig:model}
  \vspace{-0.4cm}
\end{figure}

This section introduces the \emph{\networkname{}} (\networkshort{}), a neural network for predicting future physical states shown in Figure~\ref{fig:model}.
At each time step $t$, \networkshort{} takes a history of $T$ previous particle states $P^{(t-T, t]}$ and relations $R^{(t-T, t]}$ in the form of hierarchical scene graphs $G_H^{(t-T, t]}$ as input. $G_{H}^{(t-T, t]}$ dynamically changes over time as directed, unlabeled virtual collision relations are added for sufficiently close pairs of particles. \networkshort{} also takes external effects on the system (for example gravity $g$ or external forces $F$) as input.
The model consists of three pairwise graph convolution modules, one for external forces ($\phi_F$), one for collisions ($\phi_C$) and one for past states ($\phi_H$), followed by a hierarchical graph convolution module $\eta$ that propagates effects through the particle hierarchy. A fully-connected module $\psi$ then outputs the next states $P^{t+1}$.

In the following, we briefly describe each module. For ease of reading we \emph{drop the notation ${(t-T, t]}$} and assume that all variables are subject to this time range unless otherwise noted.

\textbf{External Force Module}\ \ \ 
The \emph{external force module} $\phi_F$ converts forces $F \equiv \{f_i\}$ on leaf particles $p_i \in P^L$ into effects $\phi_F(p_i, f_i) = e^F_i \in \R^E$. 

\textbf{Collision Module}\ \ \ 
Collisions between objects are handled by dynamically defining pairwise collision relations $r_{ij}^C$ between leaf particles $p_i  \in P^L$ from one object and $p_j \in P^L$ from another object that are close to each other \citep{chang2016compositional}. The \emph{collision module} $\phi_C$ uses $p_i$, $p_j$ and $r_{ij}^C$ to compute the effects $\phi_C(p_j, p_i, r_{ij}^C) = e^C_{ji} \in \R^E$ of $p_j$ on $p_i$ and vice versa. With $d^{t}(i,j)=\|x^t_{i} - x^t_{j}\|$, the overall collision effects equal $e^C_i = \sum_j \{e_{ji}| d^{t}(i,j) <D_C\}$. The hyperparameter $D_C$ represents the maximum distance for a collision relation. 

\textbf{History Module}\ \ \ 
The \emph{history module} $\phi_H$ predicts the effects $\phi(p_i^{(t-T,t-1]}, p_i^{t}) \in e^H_i$ from past $p_i^{(t-T,t-1]} \in P^L$ on current leaf particle states $p_i^{t} \in P^L$.

\textbf{Hierarchical Effect Propagation Module}\ \ \ 
The \emph{hierarchical effect propagation module} $\eta$ propagates the overall effect $e^0_i = e^F_i+e^C_i+e^H_i$ from external forces, collisions and history on $p_i$ through the particle hierarchy. $\eta$ corresponds to the three-stage hierarchical graph convolution introduced in Figure~\ref{fig:graphconv}~b) which given the pairwise particle states $p_i$ and $p_j$, their relation $r_{ij}$, and input effects $e^0_i$, outputs the total propagated effect $e_i$ on each particle $p_i$.

\textbf{State Prediction Module}\ \ \ 
We use a simple fully-connected network $\psi$ to predict the next particle states $P^{t+1}$. In order to get more accurate predictions, we leverage the hierarchical particle representation by predicting the dynamics of any given particle within the local coordinate system originated at its parent. The only exceptions are object root particles for which we predict the global dynamics. 
Specifically, the \emph{state prediction module} $\psi(g, p_i, e_{i})$ predicts the local future delta position $\delta^{t+1}_{i,\ell} = \delta^{t+1}_{i} - \delta^{t+1}_{\textrm{par}(i)}$ using the particle state $p_i$, the total effect $e_{i}$ on $p_i$, and the gravity $g$ as input. As we only predict global dynamics for object root particles, the gravity is only applied to these root particles. The final future delta position in world coordinates is computed from local information as $\delta^{t+1}_{i}=\delta^{t+1}_{i,\ell} + \sum_j\delta^{t+1}_{j, \ell}, j\in \textrm{anc}(i)$.




\subsection{Learning Physical Constraints through Loss Functions and Data}
Traditionally, physical systems are modeled with equations providing fixed approximations of the real world. Instead, we choose to learn physical constraints, including the meaning of the material property vector, from data. The error signal we found to work best is a combination of three objectives. (1) We predict the position change $\delta^{t+1}_{i, \ell}$ between time step $t$ and $t+1$ independently for all particles in the hierarchy. 
In practice, we find that $\delta^{t+1}_{i,\ell}$ will differ in magnitude for particles in different levels. 
Therefore, we normalize the local dynamics using the statistics from all particles in the same level (\emph{local loss}).
(2) We also require that the global future delta position $\delta^{t+1}_{i}$ is accurate (\emph{global loss}).
(3) We aim to preserve the intra-object particle structure by imposing that the pairwise distance between two connected particles $p_i$ and $p_j$ in the next time step $d^{t+1}(i,j)$ matches the ground truth. In the case of a rigid body this term works to preserve the distance between particles. For soft bodies, this objective ensures that pairwise local deformations are learned correctly (\emph{preservation loss}).

The total objective function linearly combines (1), (2), and (3) weighted by hyperparameters $\alpha$ and $\beta$:
\begin{equation*}
\textit{Loss} = \alpha \bigr (\sum_{p_i} \|\hat \delta^{t+1}_{i, \ell} - \delta^{t + 1}_{i, \ell} \|^2 + 
\beta \sum_{p_i} \|\hat \delta^{t+1}_{i} - \delta^{t + 1}_{i} \|^2 \bigr) +  \bigr( 1-\alpha \bigr) \sum_{p_i \in \textrm{sib}(p_j)}
{\|\hat{d}^{t+1}(i,j) - d^{t+1}(i,j)\|^2}
\end{equation*}


\section{Experiments}
In this section, we examine the \networkshort{}'s ability to accurately predict the physical state across time in scenarios with rigid bodies, deformable bodies (soft bodies, cloths, and fluids), collisions, and external actions. We also evaluate the generalization performance across various object and environment properties. Finally, we present some more complex scenarios including (e.g.) falling block towers and dominoes. Prediction roll-outs are generated by recursively feeding back the HRN's one-step prediction as input. We strongly encourage readers to have a look at result examples shown in main text figures, supplementary materials, and at \url{https://youtu.be/kD2U6lghyUE}. 

All training data for the below experiments was generated via a custom \emph{interactive particle-based environment} based on the FleX physics engine \citep{macklin2014unified} in Unity3D. This environment provides (1) an automated way to extract a particle representation given a 3D object mesh, (2) a convenient way to generate randomized physics scenes for generating static training data, and (3) a standardized way to interact with objects in the environment through forces.\footnote[2]{HRN code and Unity FleX environment can be found at \url{https://neuroailab.github.io/physics/}}. Further details about the experimental setups and training procedure can be found in the supplement.


\begin{figure}[!ht]
  \centering
  \includegraphics[width=\textwidth]{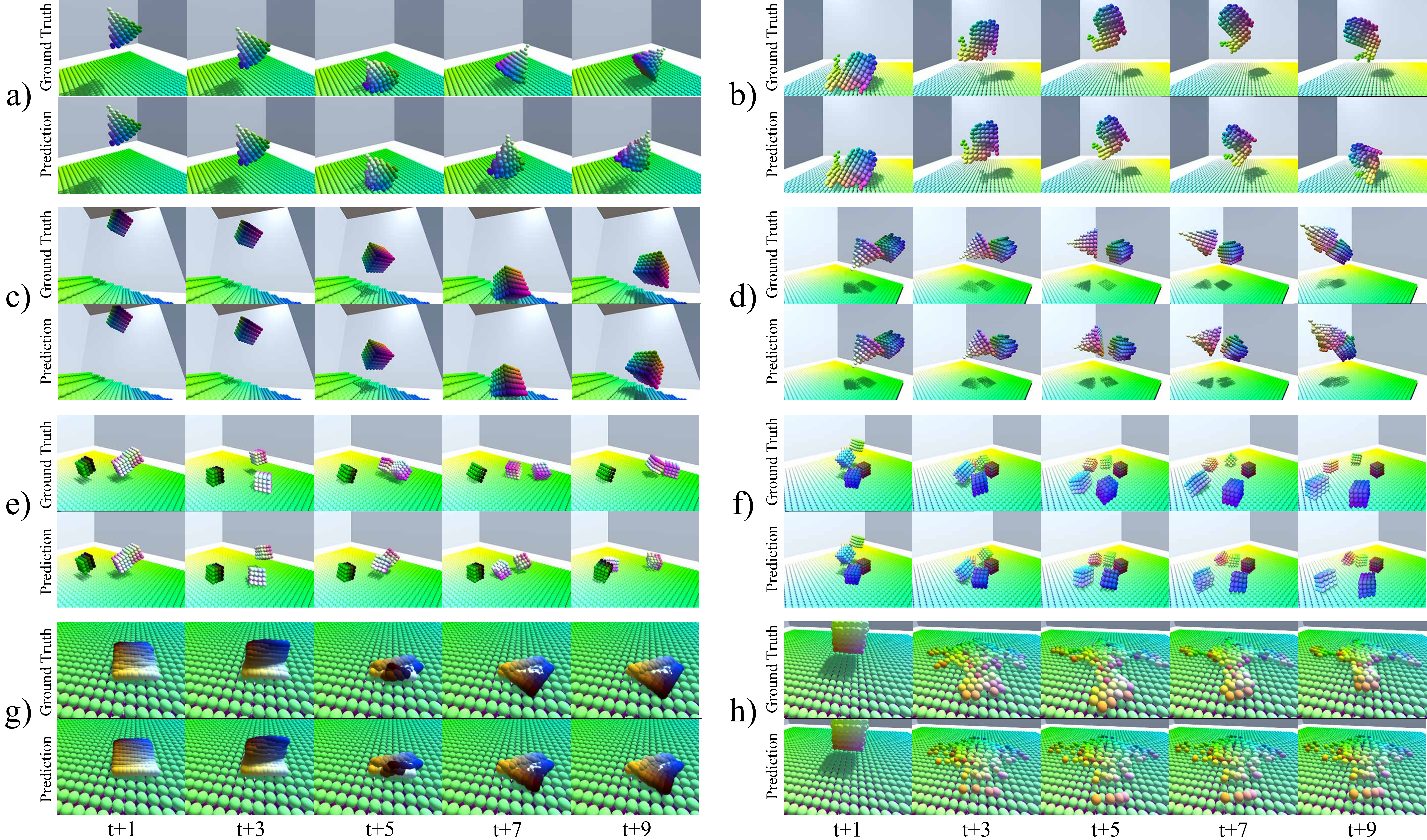}
  \caption{\textbf{Prediction examples and ground truth.}
  \textbf{a)} A cone bouncing off a plane. 
  \textbf{b)} Parabolic motion of a bunny. 
  A force is applied at the first frame.
  \textbf{c)} A cube falling on a slope.
  \textbf{d)} A cone colliding with a pentagonal prism. 
  Both shapes were held-out. 
  \textbf{e)} Three objects colliding on a plane.
  \textbf{f)} Falling block tower not trained on. 
  \textbf{g)} A cloth drops and folds after hitting the floor. 
  \textbf{h)} A fluid drop bursts on the ground. We strongly recommend watching the videos in the supplement.}
  \label{fig:pred_vis}
  \vspace{-0.6cm}
\end{figure}

\subsection{Qualitative evaluation of physical phenomena}

\textbf{\emph{Rigid body kinematic motion and external forces.}}
In a first experiment, rigid objects are pushed up, via an externally applied force, from a ground plane then fall back down and collide with the plane. The model is trained on 10 different simple shapes (cube, sphere, pyramid, cylinder, cuboid, torus, prism, octahedron, ellipsoid, flat pyramid) with 50-300 particles each. The static plane is represented using 5,000 particles with a practically infinite mass.
External forces spatially dispersed with a Gaussian kernel are applied at randomly chosen points on the object.
Testing is performed on instances of the same rigid shapes, but with new force vectors and application points, resulting in new trajectories. 
Results can be seen in supplementary Figure~\ref{fig:more_examples}c-d, illustrating that the \networkshort{} correctly predicts the parabolic kinematic trajectories of tangentially accelerated objects, rotation due to torque, responses to initial external impulses, and the eventual elastic collisions of the object with the floor. 

\textbf{\emph{Complex shapes and surfaces.}} In more complex scenarios, we train on the simple shapes colliding with a plane then generalize to complex non-convex shapes (e.g. bunny, duck, teddy). Figure~\ref{fig:pred_vis}b shows an example prediction for the bunny; more examples are shown in supplementary Figure~\ref{fig:more_examples}g-h. 

We also examine spheres and cubes falling on 5 complex surfaces: slope, stairs, half-pipe, bowl, and a ``random'' bumpy surface. See Figure~\ref{fig:pred_vis}c and supplementary Figure~\ref{fig:more_examples2}c-e for results. We train on spheres and cubes falling on the 5 surfaces, and test on new trajectories. 

\textbf{\emph{Dynamic collisions.}}
Collisions between two moving objects are more complicated to predict than static collisions (e.g. between an object and the ground). We first evaluate this setup in a zero-gravity environment to obtain purely dynamic collisions. Training was performed on collisions between 9 pairs of shapes sampled from the 10 shapes in the first experiment. 
Figure~\ref{fig:pred_vis}d shows predictions for collisions involving shapes not seen during training, the cone and pentagonal prism, demonstrating HRN's ability to generalize across shapes. Additional examples can be found in supplementary Figure~\ref{fig:more_examples}e-f, showing results on trained shapes.

\textbf{\emph{Many-object interactions.}}
Complex scenarios include simultaneous interactions between multiple moving objects supported by static surfaces. For example, when three objects collide on a planar surface, the model has to resolve direct object collisions, indirect collisions through intermediate objects, and forces exerted by the surface to support the objects. To illustrate the HRN's ability to handle such scenarios, we train on combinations of two and three objects (cube, stick, sphere, ellipsoid, triangular prism, cuboid, torus, pyramid) colliding simultaneously on a plane. 
See Figure~\ref{fig:pred_vis}e and supplementary Figure~\ref{fig:more_examples2}f for results. 

We also show that HRN trained on the two and three object collision data generalizes to complex new scenarios. Generalization tests were performed on a falling block tower, a falling domino chain, and a bowl containing multiple spheres. All setups consist of 5 objects. See Figure~\ref{fig:pred_vis}f and supplementary Figures~\ref{fig:more_examples}b and \ref{fig:more_examples2}b,g for results. Although predictions sometimes differ from ground truth in their details, results still appear plausible to human observers. 

\textbf{\emph{Soft bodies.}} We repeat the same experiments but with soft bodies of varying stiffness, showing that \networkshort{} properly handles kinematics, external forces, and collisions with complex shapes and surfaces involving soft bodies. One illustrative result is depicted in Figure \ref{fig:setup}, showing a non-rigid cube as it deformably bounces off the floor.  Additional examples are shown in supplementary Figure~\ref{fig:more_examples}g-h.

\textbf{\emph{Cloth.}} We also experiment with various cloth setups. In the first experiment, a cloth drops on the floor from a certain height and folds or deforms. In another experiment a cloth is fixated at two points and swings back and forth. Cloth predictions are very challenging as cloths do not spring back to their original shape and self-collisions have to be resolved in addition to collisions with the ground. To address this challenge, we add self-collisions, collision relationships between particles within the same object, in the collision module. Results can be seen in Figure~\ref{fig:pred_vis}g and supplementary Figure~\ref{fig:cloth} and show that the cloth motion and deformations are accurately predicted.

\textbf{\emph{Fluids.}} In order to test our models ability to predict fluids, we perform a simple experiment in which a fluid drop drops on the floor from a certain height. As effects within a fluid are mostly local, flat hierarchies with small groupings are better on fluid prediction. Results can be seen in Figure~\ref{fig:pred_vis}h and show that the fall of a liquid drop is successfully predicted when trained in this scenario.


\textbf{\emph{Response to parameter variation.}}
To evaluate how the HRN responds to changes in mass, gravity and stiffness, we train on datasets in which these properties vary. During testing time we vary those parameters for the same initial starting state and evaluate how trajectories change. In supplementary Figures \ref{fig:var_mass}, \ref{fig:var_grav} and \ref{fig:var_stiff} we show results for each variation, illustrating e.g. how objects accelerate more rapidly in a stronger gravitational field. 

\textbf{\emph{Heterogeneous materials.}}
We leverage the hierarchical particle graph representation to construct objects that contain both rigid and soft parts. After training a model with objects of varying shapes and stiffnesses falling on a plane, we manually adjust individual stiffness relations to create a half-rigid half-soft object and generate HRN predictions. Supplementary Figure \ref{fig:more_examples2}h shows a half-rigid half-soft pyramid. Note that there is no ground truth for this example as we surpass the capabilities of the used physics simulator which is incapable of simulating objects with heterogeneous materials.

\subsection{Quantitative evaluation and ablation}
We compare \networkshort{} to several baselines and model ablations. The first baseline is a simple Multi-Layer-Perceptron (MLP) which takes the full particle representation and directly outputs the next particle states. The second baseline is the Interaction Network as defined by \citet{battaglia2016interaction} denoted as \emph{fully connected graph} as it corresponds to removing our hierarchy and computing on a fully connected graph.  
In addition, to show the importance of the $\phi_C$, $\phi_F$, and $\phi_H$ modules, we remove and replace them with simple alternatives.
\emph{No $\phi_F$} replaces the force module by concatenating the forces to the particle states and directly feeding them into $\eta$. Similarly for \emph{no $\phi_C$}, $\phi_C$ is removed by adding the collision relations to the object relations and feeding them directly through $\eta$. In case of \emph{no $\phi_H$}, $\phi_H$ is simply removed and not replaced with anything.
Next, we show that two input time steps $(t,t-1)$ improve results by comparing it with a \emph{1 time step} model.
Lastly, we evaluate the importance of the \emph{preservation loss} and the \emph{global loss} component added to the \emph{local loss}. All models are trained on scenarios where two cubes collide fall on a plane and repeatedly collide after being pushed towards each other. The models are tested on held-out trajectories of the same scenario. An additional evaluation of different grouping methods can be found in Section \ref{supp:grouping} of the supplement.

\begin{figure}[!ht]
  \centering
  \vspace{-0.2cm}
  \includegraphics[width=\textwidth]{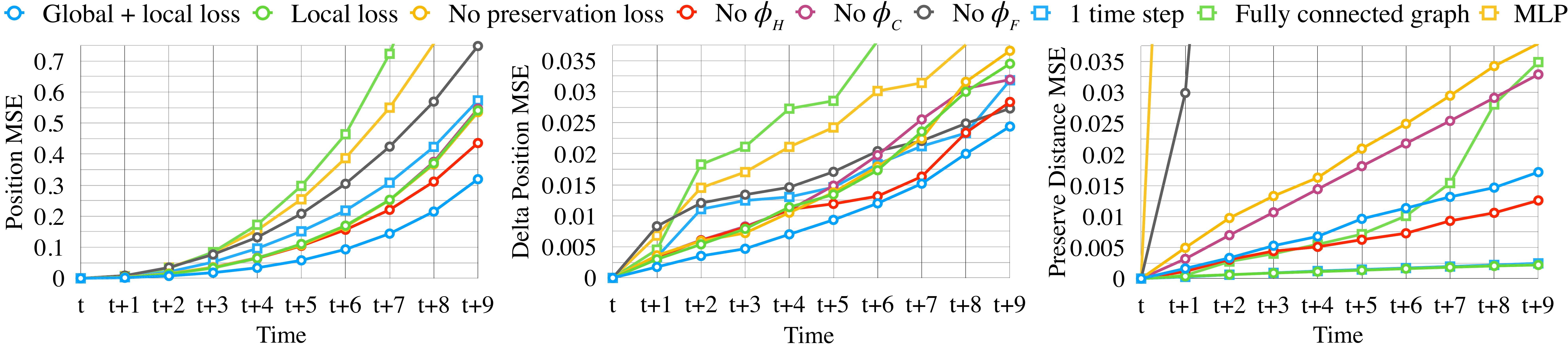}
  \caption{\textbf{Quantitative evaluation.} We compare the full \emph{HRN (global + local loss)} to several baselines, namely \emph{local loss only}, \emph{no preservation loss}, \emph{no $\phi_H$}, \emph{no $\phi_C$}, \emph{no $\phi_F$}, \emph{1 time step}, \emph{fully connected graph} and a \emph{MLP} baseline. The line graphs from left to right show the mean squared error (MSE) between positions, delta positions and distance preservation accumulated over time. Our model has the lowest position and delta position error and a only slightly higher preservation error.}
  \label{fig:msetable}
  \vspace{-0.2cm}
\end{figure}

Comparison metrics are the cumulative mean squared error of the absolute global position, local position delta, and preserve distance error up to time step $t+9$. Results are reported in Figure \ref{fig:msetable}. The \networkshort{} outperforms all controls most of the time. The hierarchy is especially important, with the \emph{fully connected graph} and \emph{MLP} baselines performing substantially worse. Besides, the \networkshort{} without the hierarchical graph convolution mechanism performed significantly worse as seen in supplementary Figure \ref{fig:quantitative_losses}, which shows the necessity of the three consecutive graph convolution stages. In qualitative evaluations, we found that using more than one input time step improves results especially during collisions as the acceleration is better estimated which the metrics in Figure \ref{fig:msetable} confirm. We also found that splitting collisions, forces, history and effect propagation into separate modules with separate weights allows each module to specialize, improving predictions. Lastly, the proposed loss structure is crucial to model training. Without distance preservation or the global delta position prediction our model performs much worse. See supplementary Section \ref{supp:lossgraph} for further discussion on the losses and graph structures.

\subsection{Discussion}
Our results show that the vast majority of complex multi-object interactions are predicted well, including multi-point collisions between non-convex geometries and complex scenarios like the bowl containing multiple rolling balls. Although not shown, in theory, one could also simulate shattering objects by removing enough relations between particles within an object. These manipulations are of substantial interest because they go beyond what is possible to generate in our simulation environment. Additionally, predictions of especially challenging situations such as multi-block towers were also mostly effective, with objects (mostly) retaining their shapes and rolling over each other convincingly as towers collapsed (see the supplement and the video). The loss of shape preservation over time can be partially attributed to the compounding errors generated by the recursive roll-outs.
Nevertheless, our model predicts the tower to collapse faster than ground truth. Predictions also jitter when objects should stand absolutely still.
These failures are mainly due to the fact that the training set contained only interactions between fast-moving pairs or triplets of objects, with no scenarios with objects at rest.
That it generalized to towers as well as it did is a powerful illustration of our approach. Adding a fraction of training observations with objects at rest causes towers to behave more realistically and removes the jitter overall. The training data plays a crucial role in reaching the final model performance and its generalization ability. Ideally, the training set would cover the entirety of physical phenomena in the world. However, designing such a dataset by hand is intractable and almost impossible. Thus, methods in which a self-driven agent sets up its own physical experiments will be crucial to maximize learning and understanding\cite{haber2018learning}. 

\section{Conclusion}
We have described a hierarchical graph-based scene representation that allows the scalable specification of arbitrary geometrical shapes and a wide variety of material properties. Using this representation, we introduced a learnable neural network based on hierarchical graph convolution that generates plausible trajectories for complex physical interactions over extended time horizons, generalizing well across shapes, masses, external and internal forces as well as material properties.
Because of the particle-based nature of our representation, it naturally captures object permanence identified in cognitive science as a key feature of human object perception~\citep{spelke1992origins}. 

A wide variety of applications of this work are possible. Several of interest include developing predictive models for grasping of rigid and soft objects in robotics, and modeling the physics of 3D point cloud scans for video games or other simulations.
To enable a pixel-based end-to-end trainable version of the HRN for use in key computer vision applications, it will be critical to combine our work with adaptations of existing methods (e.g. \citep{wu2016single, kipf2018neural,fan2017point}) for inferring initial (non-hierarchical) scene graphs from LIDAR/RGBD/RGB image or video data.
In the future, we also plan to remedy some of HRN's limitations, expanding the classes of materials it can handle to including inflatables or gases, and to dynamic scenarios in which objects can shatter or merge. This should involve a more sophisticated representation of material properties as well as a more nuanced hierarchical construction.
Finally, it will be of great interest to evaluate to what extent HRN-type models describe patterns of human intuitive physical knowledge observed by cognitive scientists~\citep{mccloskey1980curvilinear,piloto2018probing,riochet2018intphys}.

\subsubsection*{Acknowledgments}
We thank Viktor Reutskyy, Miles Macklin, Mike Skolones and Rev Lebaredian for helpful discussions and their support with integrating NVIDIA FleX into our simulation environment. This work was supported by grants from the James S. McDonnell Foundation, Simons Foundation, and Sloan Foundation (DLKY), a Berry Foundation postdoctoral fellowship (NH), the NVIDIA Corporation, ONR - MURI (Stanford Lead) N00014-16-1-2127 and ONR - MURI (UCLA Lead) 1015 G TA275.

\bibliographystyle{abbrvnat}
\bibliography{references}

\begin{thebibliography}{54}
\providecommand{\natexlab}[1]{#1}
\providecommand{\url}[1]{\texttt{#1}}
\expandafter\ifx\csname urlstyle\endcsname\relax
  \providecommand{\doi}[1]{doi: #1}\else
  \providecommand{\doi}{doi: \begingroup \urlstyle{rm}\Url}\fi

\bibitem[Agrawal et~al.(2016)Agrawal, Nair, Abbeel, Malik, and
  Levine]{agrawal2016learning}
P.~Agrawal, A.~V. Nair, P.~Abbeel, J.~Malik, and S.~Levine.
\newblock Learning to poke by poking: Experiential learning of intuitive
  physics.
\newblock In \emph{Advances in Neural Information Processing Systems}, pages
  5074--5082, 2016.

\bibitem[Baraff(2001)]{baraff2001physically}
D.~Baraff.
\newblock Physically based modeling: Rigid body simulation.
\newblock \emph{SIGGRAPH Course Notes, ACM SIGGRAPH}, 2\penalty0 (1):\penalty0
  2--1, 2001.

\bibitem[Bates et~al.(2015)Bates, Battaglia, Yildirim, and
  Tenenbaum]{bates2015humans}
C.~Bates, P.~Battaglia, I.~Yildirim, and J.~B. Tenenbaum.
\newblock Humans predict liquid dynamics using probabilistic simulation.
\newblock In \emph{CogSci}, 2015.

\bibitem[Battaglia et~al.(2016)Battaglia, Pascanu, Lai, Jimenez~Rezende, and
  kavukcuoglu]{battaglia2016interaction}
P.~Battaglia, R.~Pascanu, M.~Lai, D.~Jimenez~Rezende, and k.~kavukcuoglu.
\newblock Interaction networks for learning about objects, relations and
  physics.
\newblock In \emph{Advances in Neural Information Processing Systems 29}, pages
  4502--4510. 2016.

\bibitem[Battaglia et~al.(2013)Battaglia, Hamrick, and
  Tenenbaum]{battaglia2013simulation}
P.~W. Battaglia, J.~B. Hamrick, and J.~B. Tenenbaum.
\newblock Simulation as an engine of physical scene understanding.
\newblock \emph{Proceedings of the National Academy of Sciences}, 110\penalty0
  (45):\penalty0 18327--18332, 2013.

\bibitem[Bender et~al.(2015)Bender, M{\"u}ller, and
  Macklin]{bender2015position}
J.~Bender, M.~M{\"u}ller, and M.~Macklin.
\newblock Position-based simulation methods in computer graphics.
\newblock In \emph{Eurographics (Tutorials)}, 2015.

\bibitem[Brand(1997)]{brand1997physics}
M.~Brand.
\newblock Physics-based visual understanding.
\newblock \emph{Computer Vision and Image Understanding}, 65\penalty0
  (2):\penalty0 192--205, 1997.

\bibitem[Bronstein et~al.(2017)Bronstein, Bruna, LeCun, Szlam, and
  Vandergheynst]{bronstein2017geometric}
M.~M. Bronstein, J.~Bruna, Y.~LeCun, A.~Szlam, and P.~Vandergheynst.
\newblock Geometric deep learning: going beyond euclidean data.
\newblock \emph{IEEE Signal Processing Magazine}, 34\penalty0 (4):\penalty0
  18--42, 2017.

\bibitem[Bruna et~al.(2013)Bruna, Zaremba, Szlam, and LeCun]{bruna2013spectral}
J.~Bruna, W.~Zaremba, A.~Szlam, and Y.~LeCun.
\newblock Spectral networks and locally connected networks on graphs.
\newblock \emph{arXiv preprint arXiv:1312.6203}, 2013.

\bibitem[Byravan and Fox(2017)]{byravan2017se3}
A.~Byravan and D.~Fox.
\newblock Se3-nets: Learning rigid body motion using deep neural networks.
\newblock In \emph{Robotics and Automation (ICRA), 2017 IEEE International
  Conference on}, pages 173--180. IEEE, 2017.

\bibitem[Chang et~al.(2016)Chang, Ullman, Torralba, and
  Tenenbaum]{chang2016compositional}
M.~B. Chang, T.~Ullman, A.~Torralba, and J.~B. Tenenbaum.
\newblock A compositional object-based approach to learning physical dynamics.
\newblock \emph{arXiv preprint arXiv:1612.00341}, 2016.

\bibitem[Coumans(2010)]{coumans2010bullet}
E.~Coumans.
\newblock Bullet physics engine.
\newblock \emph{Open Source Software: http://bulletphysics. org}, 1:\penalty0
  3, 2010.

\bibitem[Defferrard et~al.(2016)Defferrard, Bresson, and
  Vandergheynst]{defferrard2016convolutional}
M.~Defferrard, X.~Bresson, and P.~Vandergheynst.
\newblock Convolutional neural networks on graphs with fast localized spectral
  filtering.
\newblock In \emph{Advances in Neural Information Processing Systems}, pages
  3844--3852, 2016.

\bibitem[Duvenaud et~al.(2015)Duvenaud, Maclaurin, Iparraguirre, Bombarell,
  Hirzel, Aspuru-Guzik, and Adams]{duvenaud2015convolutional}
D.~K. Duvenaud, D.~Maclaurin, J.~Iparraguirre, R.~Bombarell, T.~Hirzel,
  A.~Aspuru-Guzik, and R.~P. Adams.
\newblock Convolutional networks on graphs for learning molecular fingerprints.
\newblock In \emph{Advances in neural information processing systems}, pages
  2224--2232, 2015.

\bibitem[Fan et~al.(2017)Fan, Su, and Guibas]{fan2017point}
H.~Fan, H.~Su, and L.~J. Guibas.
\newblock A point set generation network for 3d object reconstruction from a
  single image.
\newblock In \emph{CVPR}, volume~2, page~6, 2017.

\bibitem[Finn et~al.(2016)Finn, Goodfellow, and Levine]{finn2016unsupervised}
C.~Finn, I.~Goodfellow, and S.~Levine.
\newblock Unsupervised learning for physical interaction through video
  prediction.
\newblock In \emph{Advances in neural information processing systems}, pages
  64--72, 2016.

\bibitem[Fragkiadaki et~al.(2015)Fragkiadaki, Agrawal, Levine, and
  Malik]{fragkiadaki2015learning}
K.~Fragkiadaki, P.~Agrawal, S.~Levine, and J.~Malik.
\newblock Learning visual predictive models of physics for playing billiards.
\newblock \emph{arXiv preprint arXiv:1511.07404}, 2015.

\bibitem[Grzeszczuk et~al.(1998)Grzeszczuk, Terzopoulos, and
  Hinton]{grzeszczuk1998neuroanimator}
R.~Grzeszczuk, D.~Terzopoulos, and G.~Hinton.
\newblock Neuroanimator: Fast neural network emulation and control of
  physics-based models.
\newblock In \emph{Proceedings of the 25th annual conference on Computer
  graphics and interactive techniques}, pages 9--20. ACM, 1998.

\bibitem[Haber et~al.(2018)Haber, Mrowca, Fei-Fei, and
  Yamins]{haber2018learning}
N.~Haber, D.~Mrowca, L.~Fei-Fei, and D.~L. Yamins.
\newblock Learning to play with intrinsically-motivated self-aware agents.
\newblock \emph{arXiv preprint arXiv:1802.07442}, 2018.

\bibitem[Hamrick et~al.(2011)Hamrick, Battaglia, and
  Tenenbaum]{hamrick2011internal}
J.~Hamrick, P.~Battaglia, and J.~B. Tenenbaum.
\newblock Internal physics models guide probabilistic judgments about object
  dynamics.
\newblock In \emph{Proceedings of the 33rd annual conference of the cognitive
  science society}, pages 1545--1550. Cognitive Science Society Austin, TX,
  2011.

\bibitem[Hegarty(2004)]{hegarty2004mechanical}
M.~Hegarty.
\newblock Mechanical reasoning by mental simulation.
\newblock \emph{Trends in cognitive sciences}, 8\penalty0 (6):\penalty0
  280--285, 2004.

\bibitem[Henaff et~al.(2015)Henaff, Bruna, and LeCun]{henaff2015deep}
M.~Henaff, J.~Bruna, and Y.~LeCun.
\newblock Deep convolutional networks on graph-structured data.
\newblock \emph{arXiv preprint arXiv:1506.05163}, 2015.

\bibitem[Kipf et~al.(2018)Kipf, Fetaya, Wang, Welling, and
  Zemel]{kipf2018neural}
T.~Kipf, E.~Fetaya, K.-C. Wang, M.~Welling, and R.~Zemel.
\newblock Neural relational inference for interacting systems.
\newblock \emph{arXiv preprint arXiv:1802.04687}, 2018.

\bibitem[Kipf and Welling(2016)]{kipf2016semi}
T.~N. Kipf and M.~Welling.
\newblock Semi-supervised classification with graph convolutional networks.
\newblock \emph{arXiv preprint arXiv:1609.02907}, 2016.

\bibitem[Kulkarni et~al.(2014)Kulkarni, Mansinghka, Kohli, and
  Tenenbaum]{kulkarni2014inverse}
T.~D. Kulkarni, V.~K. Mansinghka, P.~Kohli, and J.~B. Tenenbaum.
\newblock Inverse graphics with probabilistic cad models.
\newblock \emph{arXiv preprint arXiv:1407.1339}, 2014.

\bibitem[Kulkarni et~al.(2015)Kulkarni, Whitney, Kohli, and
  Tenenbaum]{kulkarni2015deep}
T.~D. Kulkarni, W.~F. Whitney, P.~Kohli, and J.~Tenenbaum.
\newblock Deep convolutional inverse graphics network.
\newblock In \emph{Advances in Neural Information Processing Systems}, pages
  2539--2547, 2015.

\bibitem[Lake et~al.(2017)Lake, Ullman, Tenenbaum, and
  Gershman]{lake2017building}
B.~M. Lake, T.~D. Ullman, J.~B. Tenenbaum, and S.~J. Gershman.
\newblock Building machines that learn and think like people.
\newblock \emph{Behavioral and Brain Sciences}, 40, 2017.

\bibitem[Lerer et~al.(2016)Lerer, Gross, and Fergus]{lerer2016learning}
A.~Lerer, S.~Gross, and R.~Fergus.
\newblock Learning physical intuition of block towers by example.
\newblock \emph{arXiv preprint arXiv:1603.01312}, 2016.

\bibitem[Li et~al.(2016)Li, Azimi, Leonardis, and Fritz]{li2016fall}
W.~Li, S.~Azimi, A.~Leonardis, and M.~Fritz.
\newblock To fall or not to fall: A visual approach to physical stability
  prediction.
\newblock \emph{arXiv preprint arXiv:1604.00066}, 2016.

\bibitem[Li et~al.(2015)Li, Tarlow, Brockschmidt, and Zemel]{li2015gated}
Y.~Li, D.~Tarlow, M.~Brockschmidt, and R.~Zemel.
\newblock Gated graph sequence neural networks.
\newblock \emph{arXiv preprint arXiv:1511.05493}, 2015.

\bibitem[Macklin et~al.(2014)Macklin, M{\"u}ller, Chentanez, and
  Kim]{macklin2014unified}
M.~Macklin, M.~M{\"u}ller, N.~Chentanez, and T.-Y. Kim.
\newblock Unified particle physics for real-time applications.
\newblock \emph{ACM Transactions on Graphics (TOG)}, 33\penalty0 (4):\penalty0
  153, 2014.

\bibitem[McCloskey et~al.(1980)McCloskey, Caramazza, and
  Green]{mccloskey1980curvilinear}
M.~McCloskey, A.~Caramazza, and B.~Green.
\newblock Curvilinear motion in the absence of external forces: Naive beliefs
  about the motion of objects.
\newblock \emph{Science}, 210\penalty0 (4474):\penalty0 1139--1141, 1980.

\bibitem[Mottaghi et~al.(2016{\natexlab{a}})Mottaghi, Bagherinezhad, Rastegari,
  and Farhadi]{mottaghi2016newtonian}
R.~Mottaghi, H.~Bagherinezhad, M.~Rastegari, and A.~Farhadi.
\newblock Newtonian scene understanding: Unfolding the dynamics of objects in
  static images.
\newblock In \emph{Proceedings of the IEEE Conference on Computer Vision and
  Pattern Recognition}, pages 3521--3529, 2016{\natexlab{a}}.

\bibitem[Mottaghi et~al.(2016{\natexlab{b}})Mottaghi, Rastegari, Gupta, and
  Farhadi]{mottaghi2016happens}
R.~Mottaghi, M.~Rastegari, A.~Gupta, and A.~Farhadi.
\newblock “what happens if...” learning to predict the effect of forces in
  images.
\newblock In \emph{European Conference on Computer Vision}, pages 269--285.
  Springer, 2016{\natexlab{b}}.

\bibitem[Piloto et~al.(2018)Piloto, Weinstein, Ahuja, Mirza, Wayne, Amos, Hung,
  and Botvinick]{piloto2018probing}
L.~Piloto, A.~Weinstein, A.~Ahuja, M.~Mirza, G.~Wayne, D.~Amos, C.-c. Hung, and
  M.~Botvinick.
\newblock Probing physics knowledge using tools from developmental psychology.
\newblock \emph{arXiv preprint arXiv:1804.01128}, 2018.

\bibitem[Qi et~al.(2017{\natexlab{a}})Qi, Su, Mo, and Guibas]{qi2017pointnet}
C.~R. Qi, H.~Su, K.~Mo, and L.~J. Guibas.
\newblock Pointnet: Deep learning on point sets for 3d classification and
  segmentation.
\newblock \emph{Proc. Computer Vision and Pattern Recognition (CVPR), IEEE},
  1\penalty0 (2):\penalty0 4, 2017{\natexlab{a}}.

\bibitem[Qi et~al.(2017{\natexlab{b}})Qi, Yi, Su, and Guibas]{qi2017pointnet++}
C.~R. Qi, L.~Yi, H.~Su, and L.~J. Guibas.
\newblock Pointnet++: Deep hierarchical feature learning on point sets in a
  metric space.
\newblock In \emph{Advances in Neural Information Processing Systems}, pages
  5105--5114, 2017{\natexlab{b}}.

\bibitem[Riochet et~al.(2018)Riochet, Castro, Bernard, Lerer, Fergus, Izard,
  and Dupoux]{riochet2018intphys}
R.~Riochet, M.~Y. Castro, M.~Bernard, A.~Lerer, R.~Fergus, V.~Izard, and
  E.~Dupoux.
\newblock Intphys: A framework and benchmark for visual intuitive physics
  reasoning.
\newblock \emph{arXiv preprint arXiv:1803.07616}, 2018.

\bibitem[Scarselli et~al.(2009)Scarselli, Gori, Tsoi, Hagenbuchner, and
  Monfardini]{scarselli2009graph}
F.~Scarselli, M.~Gori, A.~C. Tsoi, M.~Hagenbuchner, and G.~Monfardini.
\newblock The graph neural network model.
\newblock \emph{IEEE Transactions on Neural Networks}, 20\penalty0
  (1):\penalty0 61--80, 2009.

\bibitem[Schlichtkrull et~al.(2017)Schlichtkrull, Kipf, Bloem, Berg, Titov, and
  Welling]{schlichtkrull2017modeling}
M.~Schlichtkrull, T.~N. Kipf, P.~Bloem, R.~v.~d. Berg, I.~Titov, and
  M.~Welling.
\newblock Modeling relational data with graph convolutional networks.
\newblock \emph{arXiv preprint arXiv:1703.06103}, 2017.

\bibitem[Smith and Vul(2013)]{smith2013sources}
K.~A. Smith and E.~Vul.
\newblock Sources of uncertainty in intuitive physics.
\newblock \emph{Topics in cognitive science}, 5\penalty0 (1):\penalty0
  185--199, 2013.

\bibitem[Spelke(1990)]{spelke1990principles}
E.~S. Spelke.
\newblock Principles of object perception.
\newblock \emph{Cognitive science}, 14\penalty0 (1):\penalty0 29--56, 1990.

\bibitem[Spelke et~al.(1992)Spelke, Breinlinger, Macomber, and
  Jacobson]{spelke1992origins}
E.~S. Spelke, K.~Breinlinger, J.~Macomber, and K.~Jacobson.
\newblock Origins of knowledge.
\newblock \emph{Psychological review}, 99\penalty0 (4):\penalty0 605, 1992.

\bibitem[Sutskever and Hinton(2009)]{sutskever2009using}
I.~Sutskever and G.~E. Hinton.
\newblock Using matrices to model symbolic relationship.
\newblock In \emph{Advances in Neural Information Processing Systems}, pages
  1593--1600, 2009.

\bibitem[Tenenbaum et~al.(2011)Tenenbaum, Kemp, Griffiths, and
  Goodman]{tenenbaum2011grow}
J.~B. Tenenbaum, C.~Kemp, T.~L. Griffiths, and N.~D. Goodman.
\newblock How to grow a mind: Statistics, structure, and abstraction.
\newblock \emph{science}, 331\penalty0 (6022):\penalty0 1279--1285, 2011.

\bibitem[Tran et~al.(2015)Tran, Bourdev, Fergus, Torresani, and
  Paluri]{tran2015learning}
D.~Tran, L.~Bourdev, R.~Fergus, L.~Torresani, and M.~Paluri.
\newblock Learning spatiotemporal features with 3d convolutional networks.
\newblock In \emph{Computer Vision (ICCV), 2015 IEEE International Conference
  on}, pages 4489--4497. IEEE, 2015.

\bibitem[Tran et~al.(2016)Tran, Bourdev, Fergus, Torresani, and
  Paluri]{tran2016deep}
D.~Tran, L.~Bourdev, R.~Fergus, L.~Torresani, and M.~Paluri.
\newblock Deep end2end voxel2voxel prediction.
\newblock In \emph{Computer Vision and Pattern Recognition Workshops (CVPRW),
  2016 IEEE Conference on}, pages 402--409. IEEE, 2016.

\bibitem[Ullman et~al.(2014)Ullman, Stuhlm{\"u}ller, Goodman, and
  Tenenbaum]{ullman2014learning}
T.~Ullman, A.~Stuhlm{\"u}ller, N.~Goodman, and J.~B. Tenenbaum.
\newblock Learning physics from dynamical scenes.
\newblock In \emph{Proceedings of the 36th Annual Conference of the Cognitive
  Science society}, pages 1640--1645, 2014.

\bibitem[Wang et~al.(2018)Wang, Rosa, Yang, Wang, Trigoni, and
  Markham]{wang20183d}
Z.~Wang, S.~Rosa, B.~Yang, S.~Wang, N.~Trigoni, and A.~Markham.
\newblock 3d-physnet: Learning the intuitive physics of non-rigid object
  deformations.
\newblock \emph{arXiv preprint arXiv:1805.00328}, 2018.

\bibitem[Watters et~al.(2017)Watters, Tacchetti, Weber, Pascanu, Battaglia, and
  Zoran]{watters2017visual}
N.~Watters, A.~Tacchetti, T.~Weber, R.~Pascanu, P.~Battaglia, and D.~Zoran.
\newblock Visual interaction networks.
\newblock \emph{arXiv preprint arXiv:1706.01433}, 2017.

\bibitem[Whitney et~al.(2016)Whitney, Chang, Kulkarni, and
  Tenenbaum]{whitney2016understanding}
W.~F. Whitney, M.~Chang, T.~Kulkarni, and J.~B. Tenenbaum.
\newblock Understanding visual concepts with continuation learning.
\newblock \emph{arXiv preprint arXiv:1602.06822}, 2016.

\bibitem[Wu et~al.(2015)Wu, Yildirim, Lim, Freeman, and
  Tenenbaum]{wu2015galileo}
J.~Wu, I.~Yildirim, J.~J. Lim, B.~Freeman, and J.~Tenenbaum.
\newblock Galileo: Perceiving physical object properties by integrating a
  physics engine with deep learning.
\newblock In \emph{Advances in neural information processing systems}, pages
  127--135, 2015.

\bibitem[Wu et~al.(2016{\natexlab{a}})Wu, Lim, Zhang, Tenenbaum, and
  Freeman]{wu2016physics}
J.~Wu, J.~J. Lim, H.~Zhang, J.~B. Tenenbaum, and W.~T. Freeman.
\newblock Physics 101: Learning physical object properties from unlabeled
  videos.
\newblock In \emph{BMVC}, volume~2, page~7, 2016{\natexlab{a}}.

\bibitem[Wu et~al.(2016{\natexlab{b}})Wu, Xue, Lim, Tian, Tenenbaum, Torralba,
  and Freeman]{wu2016single}
J.~Wu, T.~Xue, J.~J. Lim, Y.~Tian, J.~B. Tenenbaum, A.~Torralba, and W.~T.
  Freeman.
\newblock Single image 3d interpreter network.
\newblock In \emph{European Conference on Computer Vision}, pages 365--382.
  Springer, 2016{\natexlab{b}}.

\end{thebibliography}

\newpage

\appendix
\renewcommand\thefigure{\thesection.\arabic{figure}}    
\setcounter{figure}{0}

{\Large \textbf{Supplementary Material}}

\section{Iterative hierarchical grouping algorithm}
We describe the iterative grouping algorithm used to generate our hierarchical particle-based object representation in Algorithm \ref{fig:grouping_algorithm}:

\begin{algorithm}[H]
 \SetKwInOut{Input}{input}
 \SetKwInOut{Output}{output}
 \Input{Scene graph $G = <P, R>$ with particles $P$ and relations $R$ and target cluster size $N_C$}
 \Output{Hierarchical scene graph $G_H = <P_H, R_H>$}
 \SetKwBlock{Beginn}{beginn}{ende}
\Begin{
  Initialize $R_H = \{\}$ and $P_H = \{\}$\;
  \For{connected component (object) $o \in G$}
  {
  Initialize $R_{o} = \{ \}$ and $P_{o} = \{p_i| i \in o \}$\;
  Create root particle $p_{\textrm{root}} = <\frac{1}{|P_o|} \Sigma_{i \in o} x_i, \frac{1}{|P_o|} \Sigma_{i \in o} \delta_i, \Sigma_{i \in o} m_i>$ \;
  Connect $p_{\textrm{root}}$ to $\textrm{leaves}(p_{\textrm{root}})$ with relations \\~$R_{\textrm{A2D}} \equiv \{r_{ij} | i=\textrm{root}; p_j \in \textrm{leaves}(p_{\textrm{root}}) \}$\;
  Connect $\textrm{leaves}(p_\textrm{root})$ to $p_{\textrm{root}}$ with relations \\~$R_{\textrm{L2A}} \equiv \{r_{ij} | p_i \in \textrm{leaves}(p_{\textrm{root}}); j=\textrm{root} \}$\;
  Add relations to $R_o \equiv R_o \cup R_{A2D} \cup R_{L2A}$\;
  Initialize the particle processing queue $q = \{p_{\textrm{root}}\}$\;
  \While{$q$ not empty}{
    Get current particle $p_{\textrm{curr}} = pop(q)$\;
    Initialize processed subcomponent indexes $I_s=\{\}$\;
    \If{$\vert \textrm{leaves}(p_{\textrm{curr}}) \vert \geq N_C$}{
      Use k-means to group $\textrm{leaves}(p_{\textrm{curr}})$ into $N_C$ subcomponents $\{S_1, S_2, ..., S_{N_C}\}$\;
      \For{subcomponent $S \in \{S_1, S_2, ..., S_{N_c}\}$}
      {
      	\If{$\vert S \vert > 1$}{
          Create new root particle for subcomponent $p_s = <\frac{1}{\vert S \vert} \Sigma_{i \in S} x_i, \frac{1}{\vert S \vert} \Sigma_{i \in S} \delta_i, \Sigma_{i \in S} m_i>$ \;
          Connect all $\textrm{anc}(p_s)$ to $p_s$ with relations $R_{\textrm{A2D}}^1 \equiv \{r_{is} | p_i \in \textrm{anc}(p_s)\}$ \;
          Connect $p_s$ to all $\textrm{leaves}(p_s)$ with relations $R_{\textrm{A2D}}^2 \equiv \{r_{sj} | p_j \in \textrm{leaves}(p_s) \}$\;
          Connect all $\textrm{leaves}(p_s)$ to $p_s$ with relations $R_{\textrm{L2A}} \equiv \{r_{is} | p_i \in \textrm{leaves}(p_s)\}$\;
          Add relations to $R_{o} \equiv R_{o} \cup R_{\textrm{A2D}}^1 \cup R_{\textrm{A2D}}^2 \cup R_{\textrm{L2A}}$\;
          Add $p_s$ to $P_{o} \equiv P_{o} \cup \{ p_s \} $\;
          Add $s$ to $I_s \equiv I_s \cup \{s\}$\;
          Append $p_s$ to processing queue $q = push(p_s, q)$\;
      	} \Else{
          Add $S$ to $I_s \equiv I_s \cup S$\;
        }
      } 
   }\Else{
     Set $I_s \equiv \{i|p_i \in \textrm{leaves}(p_\textrm{curr})\}$\;
   }
   Connect all particle pairs $p_i$ and $p_j$ in $I_s$ with $R_{\textrm{WS}} \equiv \{r_{ij} | i,j \in I_s\}$ \; 
   Add $R_{\textrm{WS}}$ to $R_{o} \equiv R_{o} \cup R_{\textrm{WS}}$ \;
 }
 Add relations $R_o$ to $R_H \equiv R_H \cup R_o$ \;
 Add particles $P_o$ to $P_H \equiv P_H \cup P_o$\;
 }
 Return $G_H = <P_H, R_H>$\;
}
 \caption{Iterative hierarchical grouping algorithm.}
 \label{fig:grouping_algorithm}
\end{algorithm}

\section{Comparison of different grouping methods}
\label{supp:grouping}
\begin{figure}[!ht]
  \centering
  \includegraphics[width=\textwidth]{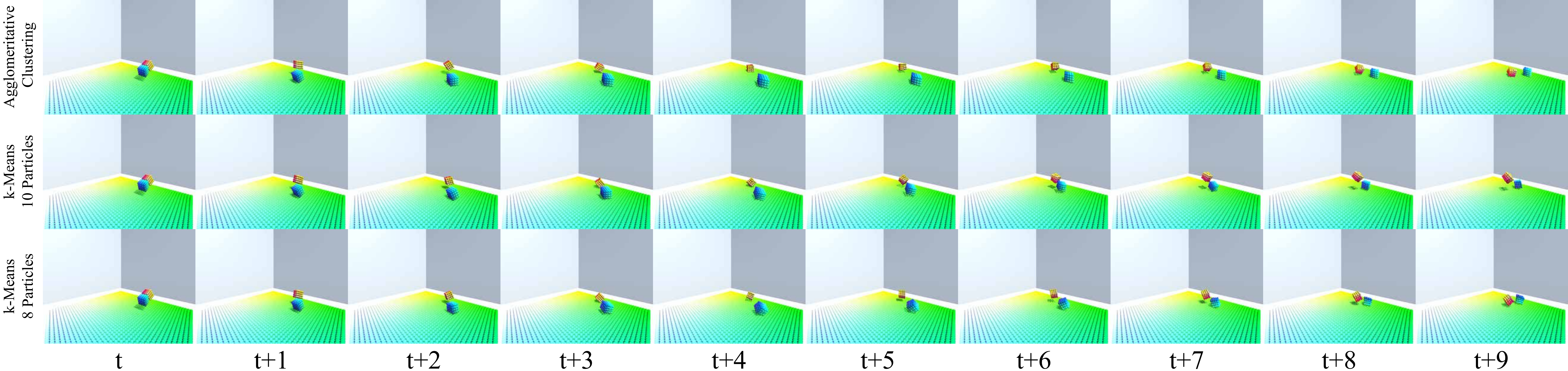}
   \caption{\textbf{Qualitative comparison of different grouping methods.} Agglomerative grouping (top) is compared against k-means with up to 10 particles per group (middle), and k-means with up to 8 particles per group (bottom) which is used in \networkshort{}.}
   \label{fig:qualitative_groupings}
\end{figure}

\begin{figure}[!ht]
  \centering
  \includegraphics[width=\textwidth]{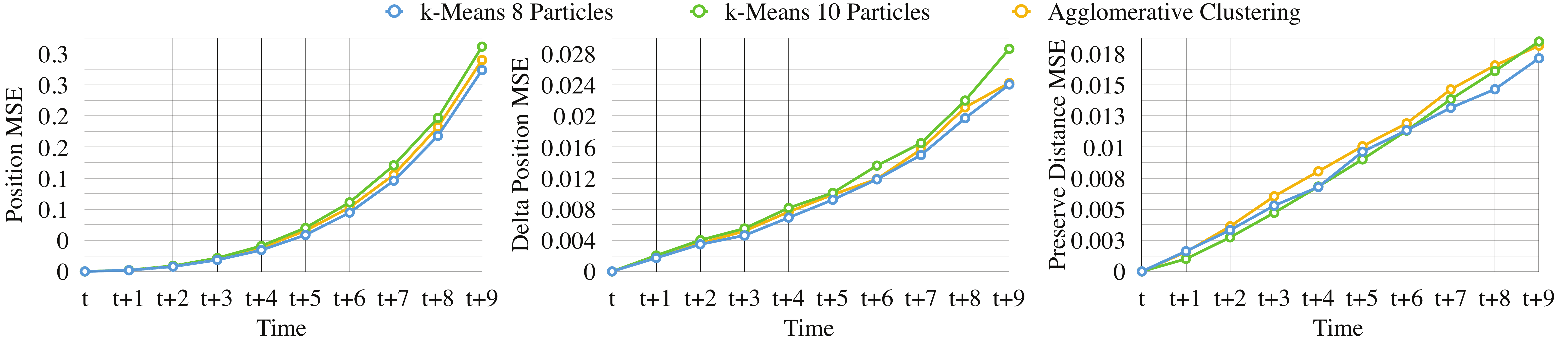}
   \caption{\textbf{Quantitative comparison of different grouping methods.} Agglomerative grouping (yellow) is compared against k-means with up to 10 particles per group (green), and k-means with up to 8 particles per group (blue) which is used in \networkshort{}.}
   \label{fig:quantitative_groupings}
\end{figure}

While performing a hyperparamter search we also tried several different grouping methods. Here, we compare agglomerative clustering against different versions of k-means. Specifically, we tried to generate hierarchies with up to 8 particles and 10 particles per group grouped by k-means. As seen in Figure \ref{fig:qualitative_groupings} and Figure \ref{fig:quantitative_groupings} we found that k-means with 8 particle groups works best resulting in a reasonable trade-off between number of particles per group and number of hierarchical layers for the tested objects. However, the improvement over the other clustering algorithms is minor, indicating that HRN is robust to the grouping method.

\section{Comparison of different losses and graph structures}
\label{supp:lossgraph}

\begin{figure}[!ht]
  \centering
  \includegraphics[width=\textwidth]{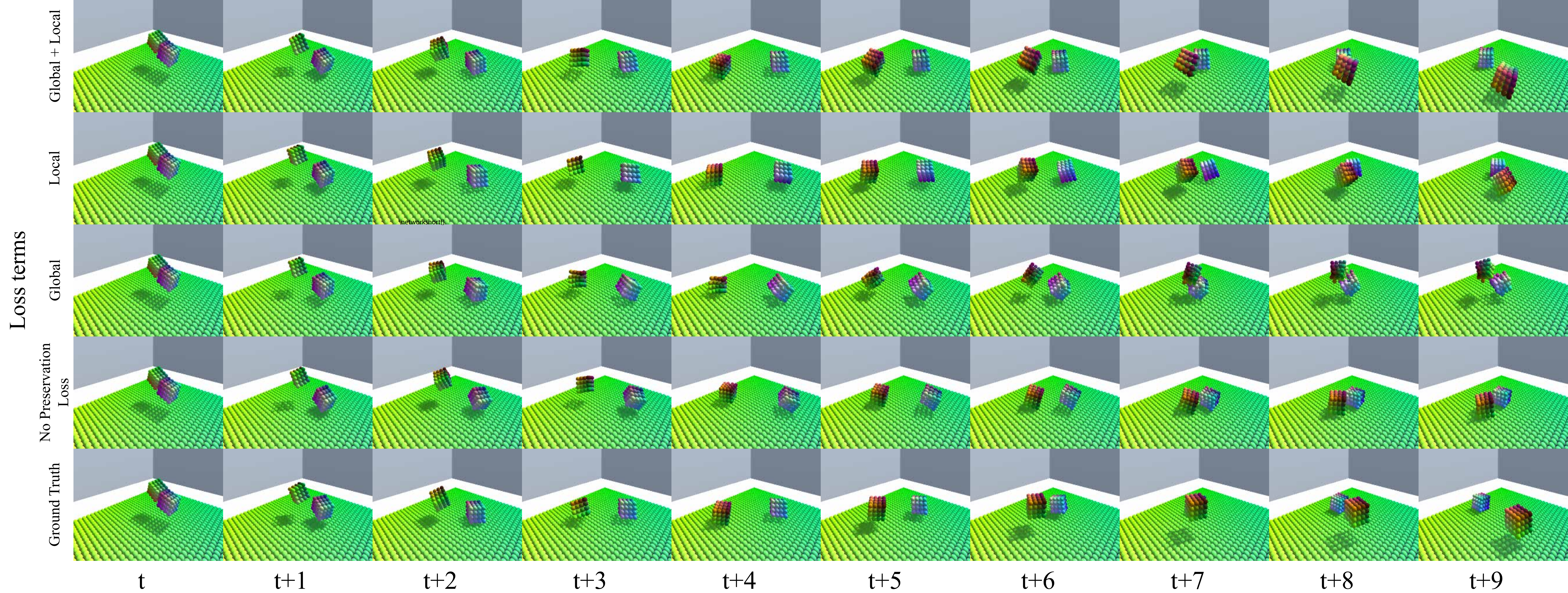}
   \caption{\textbf{Qualitative comparison of different loss terms.} Combining global and local loss terms (top) results in predictions closest to the ground truth (bottom) compared with using no preservation loss, a local loss or global loss by itself.}
   \label{fig:qualitative_losses}
\end{figure}

\begin{figure}[!ht]
  \centering
  \includegraphics[width=\textwidth]{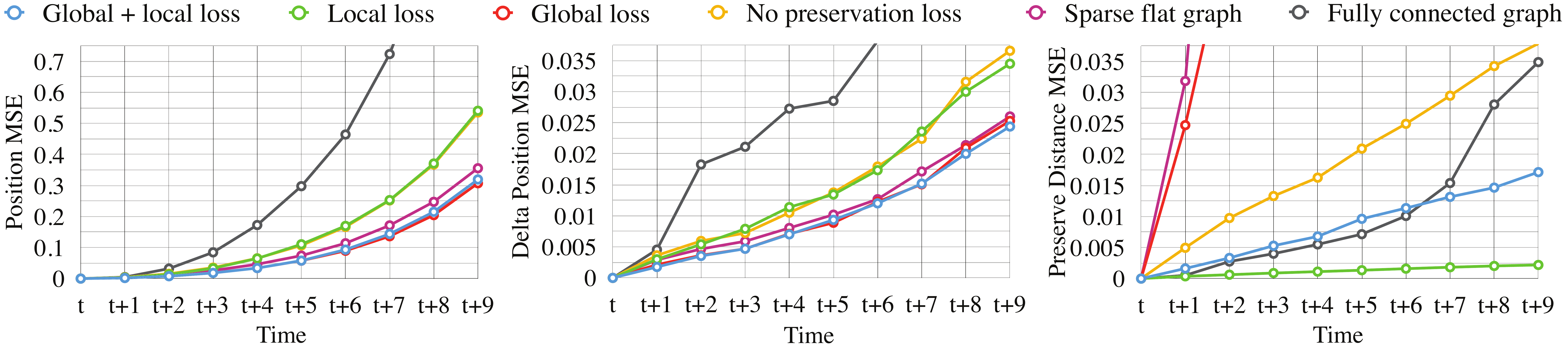}
   \caption{\textbf{Quantitative comparison of different losses and graph structures.} Losses and graph structure are ablated from left to right. In terms of losses, the full \networkshort{} with global + local loss (blue) is compared against local loss only (green), global loss only (red) and a loss without a preserve distance term (yellow). Regarding graph structure, the full \networkshort{} (blue) is compared against a sparse flat graph in which the hierarchy was removed (purple) and a fully connected graph structure (black) as presented in \citet{battaglia2016interaction}.}
   \label{fig:quantitative_losses}
\end{figure}

This section complements the quantitative evaluation and ablation studies. Figure \ref{fig:qualitative_losses} compares the predictions of models trained with the different loss terms. Results of a model trained with a combination of global and local losses are visually closest to ground truth. These qualitative results align well with the quantitative results in Figure~\ref{fig:quantitative_losses} and Figure~\ref{fig:msetable}. 

Figure \ref{fig:quantitative_losses} also illustrates the importance of a hierarchical graph (global + local loss) compared to a sparse flat graph or a fully connected graph. While the fully connected graph performs worse than the sparse flat graph and the hierarchical graph on all metrics, the sparse flat graph is comparable to the hierarchical graph on the position and delta position MSE. However, the sparse flat graph does much worse on the preserve distance MSE, indicating that the original object shape is hardly preserved. Presumably, the effect propagation in the sparse flat graph is less effective than in the hierarchical graph leading to acceptable particle positions but deformed objects.

Summarizing, a better performance on the quantitative metrics (position MSE, delta position MSE and preserve distance MSE) indeed results in qualitatively better examples. Our final combination of global and local loss terms outperforms each individual loss on its own. Similarly, our hierarchical graph significantly improves predictions compared to a sparse flat graph or a fully connected graph.

\section{Implementation details}
\subsection{Detailed model structure}
The \networkshort{} is given the states $P_o^{(t-T, t]}$, the gravity $g$ and any external forces $F$. It is trained to predict the future particle states $P_o^{t+1}$ for each object $o$.
In our implementation, the model actually predicts the change in local position $\Delta^{t+1} \equiv \mpos{}^{t+1}-\mpos{}^{t}$, and use $\Delta^{t+1}$ to advance the particle states. Note that $\mpos{}^t_o \equiv \{\lmpos{}^t_{j} | p_j \in P_{o}\}$ is the set of all particle positions in $o$.

\begin{figure}[!ht]
  \centering
  \includegraphics[width=\textwidth]{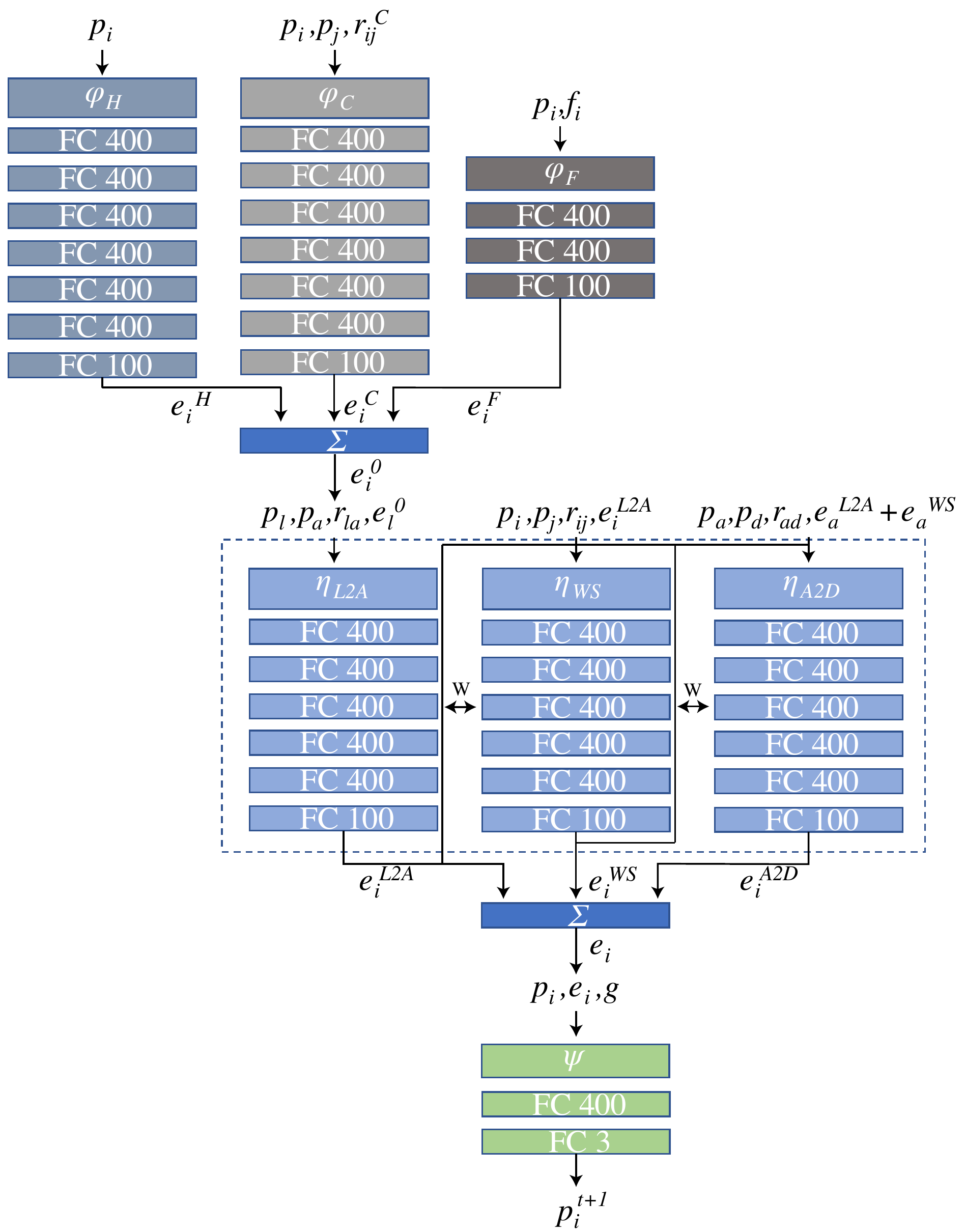}
   \caption{\textbf{Detailed description of the \networkshort{} model architecture.}}
   \label{fig:detailed_model}
\end{figure}

Figure \ref{fig:detailed_model} shows a detailed overview of \networkshort{} model architecture. In total, there are five modules, each with their own MLP. The dotted box denotes shared weights between the three hierarchical graph convolution stages, $\eta_{L2A}$, $\eta_{WS}$, and $\eta_{A2D}$. All MLPs use a ReLU nonlinearity. The number of units, layers, and output dimension of each MLP were chosen through a hyperparameter search. The gravity input $g$ to $\Psi$ is only added for the global super-particles of each object. 

\subsection{Training procedure}
We train the network using the Adam optimizer with a batch size of 256 across multiple Nvidia Titan Xp GPUs. The initial learning rate was set at 0.001 and decayed stepwise a total of 3 times, alternating between a factor of 2 and 5 each step. We used TensorFlow for the implementation. For the generalization experiments we include data augmentation in the form of random grouping, mass, and translation.

\section{Detailed experimental setups}
\label{detailed_setups}
\subsection{Particle-based physics simulation environment}

\begin{figure}[!ht]
  \centering
  \includegraphics[width=\textwidth]{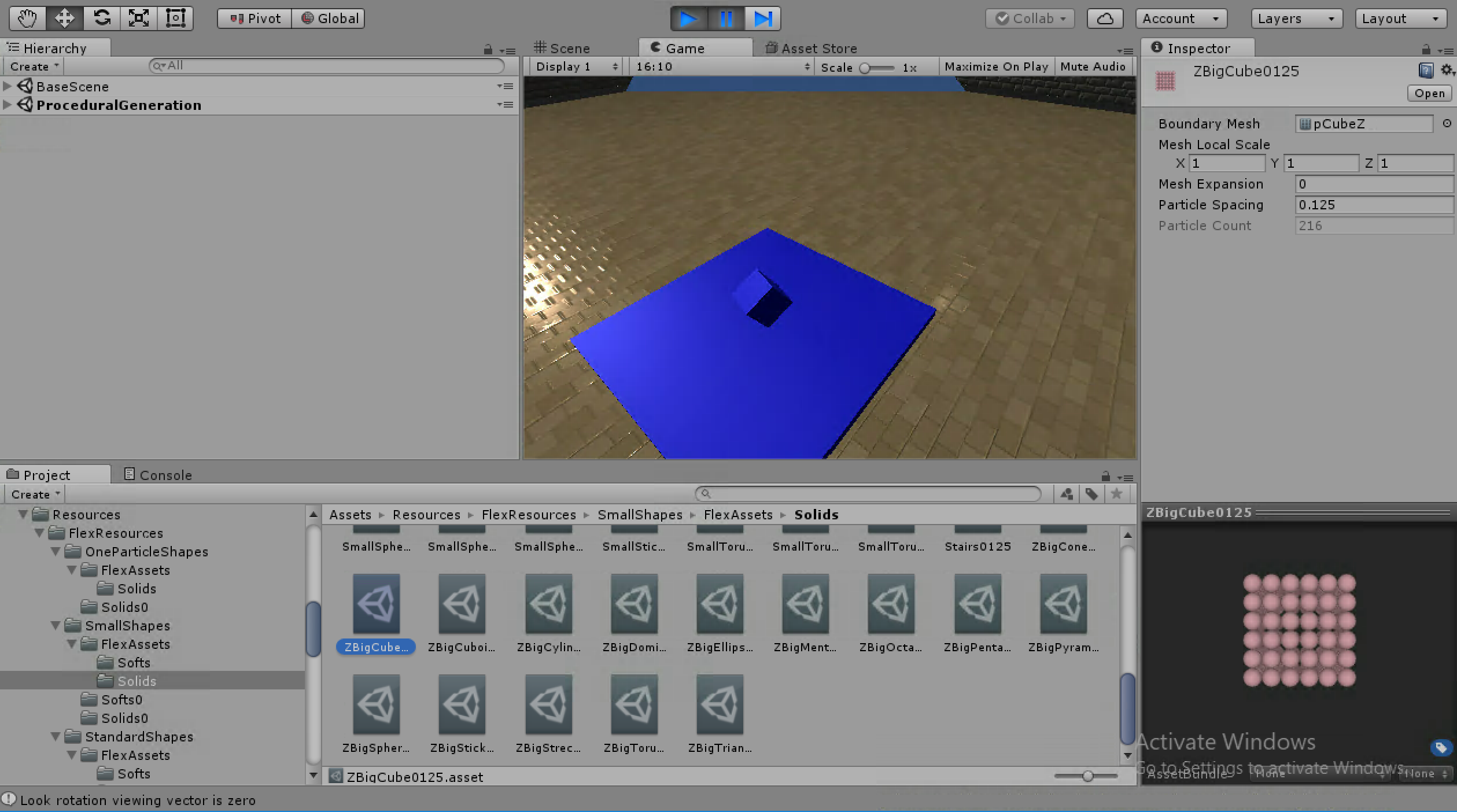}
   \caption{\textbf{Particle-based Interaction Environment in Unity3D.} Screenshot of the Unity Editor with FleX Plugin. In the main scene a cube is colliding with a planar surface. The lower right shows the particle representation of the cube. This environment is used to generate training and validation data through interactions with objects in the scene. Interactions with the environment are possible through a python interface.}
   \label{fig:unity}
\end{figure}

Based on the FleX physics engine \citep{macklin2014unified} we built a custom \emph{interactive particle-based environment} in Unity3D. This environment automatically decomposes any given 3D object mesh into a particle representation using the FleX API. On top of this representation it provides a convenient way to generate randomized physics scenes for generating static training data. The user is able to construct random scenes through a python interface that communicates with Unity3D. This interfaces also allows for physical interactions with objects within a defined scene. For instance, one can apply forces to a whole object or individual particles to generate translational and rotational position variations. It is both possible to generate static datasets from the environment and to train offline as well as to train and interact with the environment online. Therefore the environment sends the python script client the particle state at every frame as well as images captured by a camera in the scene. Scenes can be rendered with around 30 frames per second. The simulation time increases with the number of particles. Figure \ref{fig:unity} shows a screenshot of the environment embedded in the Unity3D editor. Mesh skins are used to mask the particles in the main scene to give the impression of a continuous object. In the lower right of this screenshot we can see the particle representation of the cube in the scene after FleX has converted the 3D mesh into a particle representation. Code for this environment, along with the entire HRN code base, can be found at \url{https://neuroailab.github.io/physics/}.

\subsection{Shapes and surfaces used during experiments}
\begin{figure}[!ht]
  \centering
  \includegraphics[width=\textwidth]{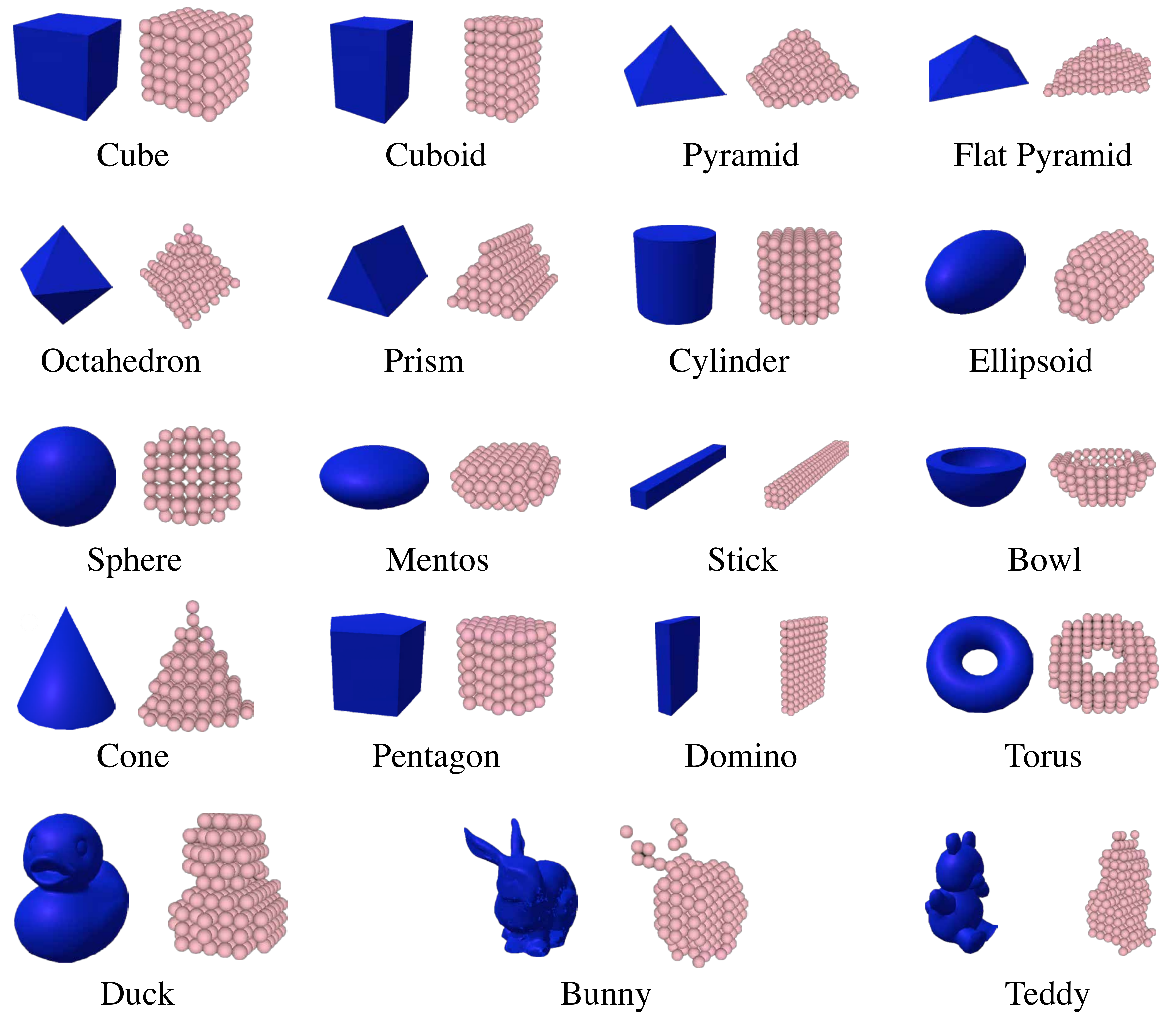}
   \caption{\textbf{Dynamic shapes and particle representations.} All shapes used during testing and training are shown. Shapes consist of 50 - 300 particles. Only one particle resolutions is shown.}
   \label{fig:shapes}
\end{figure}

\begin{figure}[!ht]
  \centering
  \includegraphics[width=\textwidth]{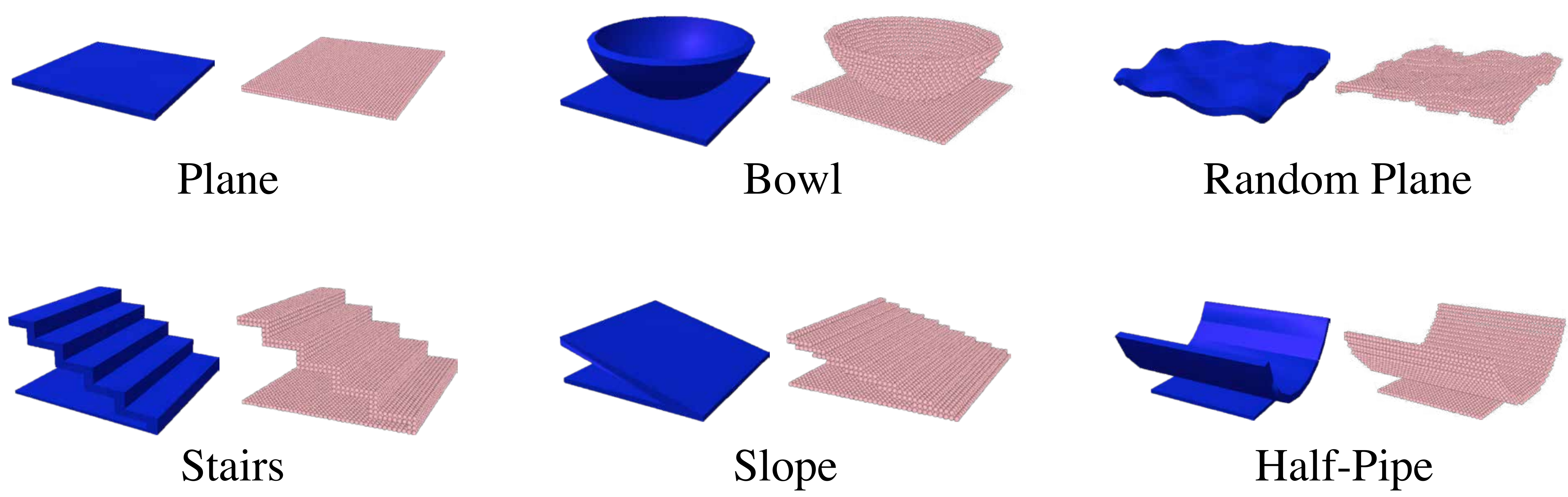}
   \caption{\textbf{Surfaces and particle representations.} All surfaces used during testing and training are shown. Surfaces consist of 5000 - 7000 particles.}
   \label{fig:surfaces}
\end{figure}

Figure \ref{fig:shapes} and Figure \ref{fig:surfaces} show the 3D mesh and the leaf particle representation of all shapes and surfaces used during training or testing. Moving objects consist of 50-300 particles, surfaces of more than 5000 particles. Only one particle resolution is shown although multiple levels of detail in the leaf node representation are possible by changing the particle spacing within an object.

\subsection{Throwing one object in the air}
In this experiment any one of the small shapes depicted in Figure \ref{fig:shapes} is first chosen to collide with one of the surfaces in Figure \ref{fig:surfaces}. The small shape is teleported to a random location around the center of the surface. The stiffness is randomly chosen per object after a teleport. As the simulation starts the shape falls on the surface and collides with it. Every random number of frames we apply a randomly upward and perpendicular to the surface pointing force to lift the object up and watch it fall again as it describes a parabola. If the object leaves the surface boundaries we randomly teleport it back to the center. After a fixed number of steps we reinitialize the scene and the whole simulation procedure starts again.

\subsection{Cloths}
Two different experiments are performed to test our model on predicting the motion of a cloth. The first experiment is similar to \emph{throwing an object in the air}. A loose cloth is teleported to a random location above the ground. On simulation start the cloth drops on the ground. Then, every fixed number of frames we apply a random force dispersed by a Gaussian kernel to the cloth and watch it deform. After a fixed number of steps we reinitialize the scene and the whole simulation procedure starts again. In the second experiments, we attach two corners of the cloth to a random location in the air. Every fixed number of steps a random force is applied to the cloth which deforms the cloth and makes it swing back and forth. The scene is reset after a fixed number of frames and the two cloth corners are attached at a new random location.

\subsection{Fluids}
In the fluid experiment a cube shaped fluid is teleported to a random location around the center of the ground. As the simulation starts the fluid drops on the ground and disperses on contact with the ground. The fluid's surface tension holds it together such that fluid particles cluster in one or few water puddles. After a set number of frames the fluid is reset to its original cube-like shape and teleported to the next random location.

\subsection{Collisions between objects without gravity}
This experimental setup is very similar to \emph{throwing an object in the air} with the difference that gravity is disabled, and we choose two small dynamic shapes that collide with each other in the air. The stiffness is randomly chosen per object after a teleport. Forces are applied such that they either point directly from one object to the other or away from each other. The force magnitude and perturbations to the force direction are randomly chosen every time an action is applied. Forces are applied randomly either to one or both objects at the same time. The simulation is reinitialized if any of the two objects leaves the room boundaries.

\subsection{Collisions between objects on a planar surface}
This experiment is a combination of the previous two experiments. Just as in \emph{throwing one object in the air} the two or three chosen small objects are spawned randomly around the center of the planar surface. The stiffness is randomly chosen per object after a teleport. They fall and collide with the plane. Similar to \emph{collisions between objects without gravity} the force is applied such that the two objects collide with each other or are torn apart. The force magnitude and perturbations to the force direction are chosen randomly. Forces are applied randomly either to one or two objects at the same time. The scene is reinitialized if any of the two objects leaves the surface boundaries.

\subsection{Stacked tower}
In this experiment we manually construct a tower consisting of 5 stacked rigid cubes on a planar surface. The positions of the cubes are slightly randomly perturbed to create towers of variable stability. After a random number of frames a force is applied to a randomly chosen cube which is usually big enough to make the tower fall. Once the tower falls and the cubes do not move anymore or after a maximum number of time steps the setup is reset and repeated.

\subsection{Dominoes}
Similar to the stacked tower, we manually setup a scene in which a rigid dominoes chain is placed on top of a planar surface. Small random perturbations are applied to the initial position of each domino. After a random number of frames a force is applied to one or both sides of the chain to make it fall. Once dominoes do not move anymore or after a fixed maximum number of time steps the setup is reset and repeated.

\subsection{Balls in bowl}
The last manually constructed control example are 5 balls dropping into a big bowl. The spheres are teleported to a randomly chosen position above the bowl. The balls then drop into the ball and interact with each other. A random force is applied every random number of frames. Once the spheres have settled or after a maximum number of time steps we reinitialize the scene.

\section{Qualitative prediction examples}
\begin{figure}[!ht]
  \centering
  \includegraphics[width=\textwidth]{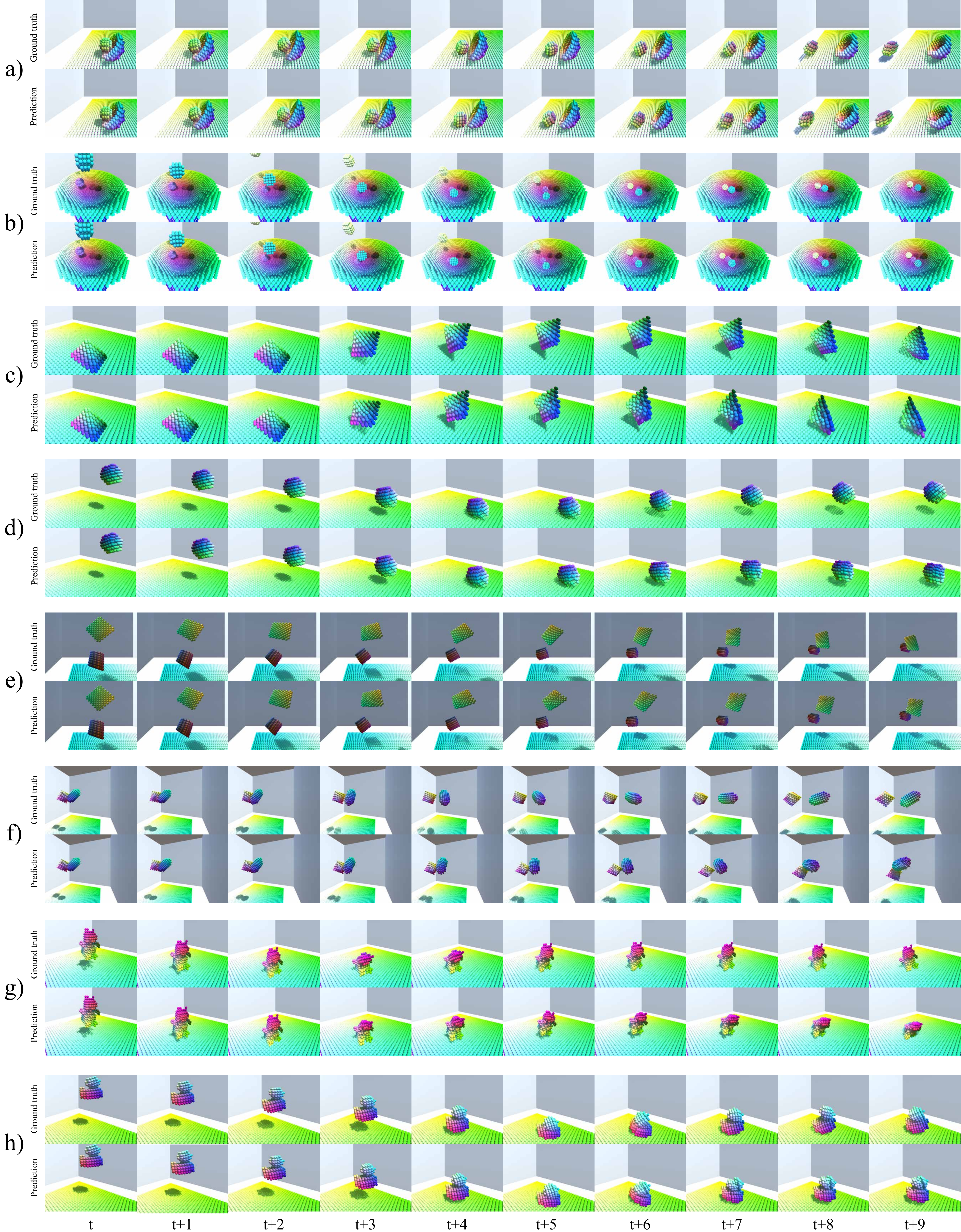}
   \caption{\textbf{Qualitative comparison of HRN predictions vs ground truth.} 
   \textbf{a)} A sphere falling out of a bowl. Objects containing other objects can be easily modeled.
   \textbf{b)} Five spheres fall into a ball and collide with each other. Complex indirect collisions occur.
   \textbf{c)} A rigid pyramid colliding with the floor.
   \textbf{d)} A rigid sphere colliding with the floor.
   \textbf{e)} A cylinder colliding with a pyramid.
   \textbf{f)} Ellipsoid and octahedron colliding with each other.
   \textbf{g)} A soft teddy colliding with the floor.
   \textbf{h)} A soft duck colliding with the floor.}
   \label{fig:more_examples}
\end{figure}

\begin{figure}[!ht]
  \centering
  \includegraphics[width=\textwidth]{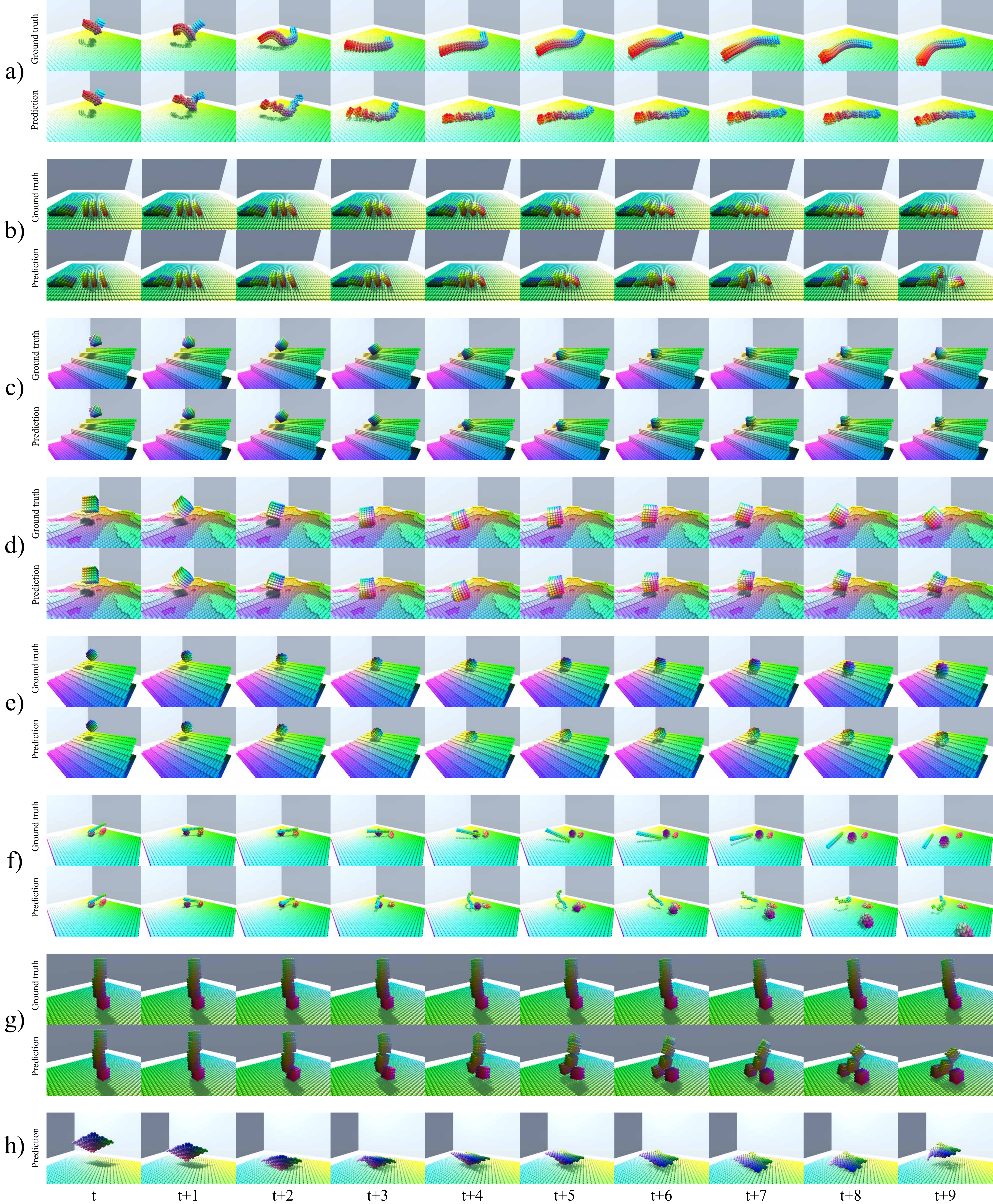}
   \caption{\textbf{Qualitative comparison of HRN predictions vs ground truth.} 
   \textbf{a)} A very deformable stick. The ground truth shape had to be fed into the model for this prediction to work.
   \textbf{b)} Falling dominoes. HRN wrongly predicts one domino moving off to the side in this complex multi-object interaction scenario.
   \textbf{c)} A rigid cube colliding with stairs.
   \textbf{d)} A cube colliding with a random surface.
   \textbf{e)} A ball on a slope.
   \textbf{f)} Three objects colliding with each other.
   \textbf{g)} A slowly falling tower. The tower in the \networkshort{} prediction collapses much faster compared to ground truth. 
   \textbf{h)} A half-rigid (right object side) half-soft (left object side) body colliding with a planar surface. The soft part deforms. The rigid part does not deform.
   }
   \label{fig:more_examples2}
\end{figure}

\begin{figure}[!ht]
  \centering
  \includegraphics[width=\textwidth]{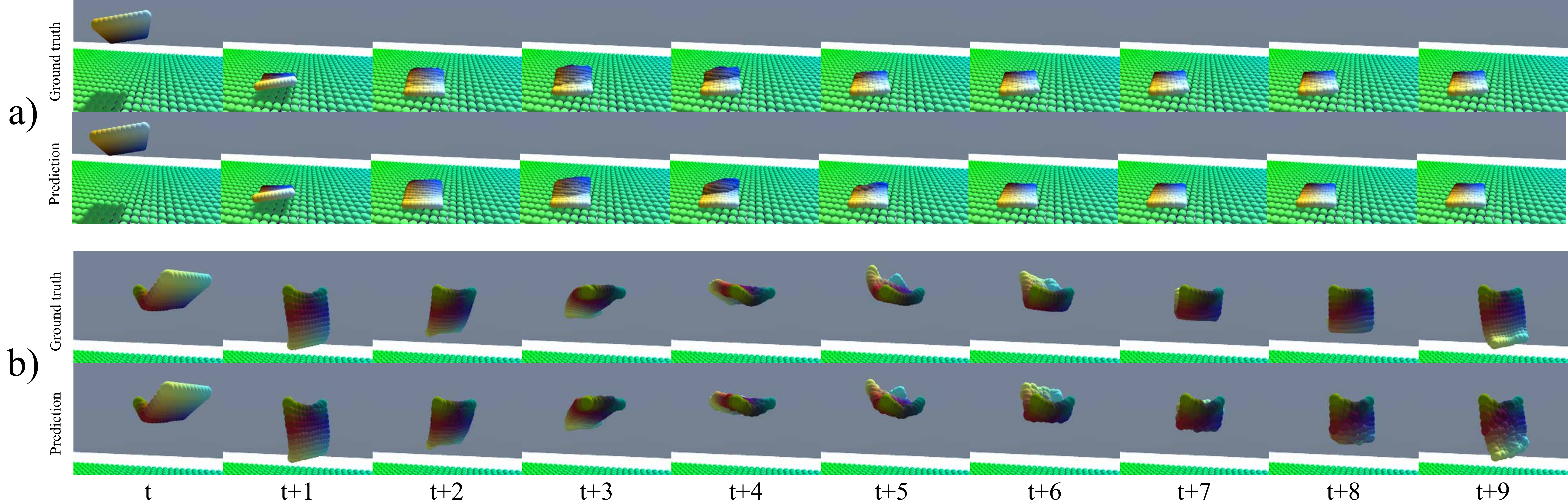}
   \caption{\textbf{Qualitative comparison of HRN predictions vs ground truth.} \textbf{a)} Dropping cloth. Cloth drops from a certain height onto the ground.
   \textbf{b)} Hanging cloth. Cloth is fixated at two points and swings back and forth.}
   \label{fig:cloth}
\end{figure}

This section showcases additional qualitative prediction examples. Figure~\ref{fig:more_examples} and Figure~\ref{fig:more_examples2} show additional examples with different objects and physical setups and failure cases. Figure~\ref{fig:cloth} visualizes additional cloth predictions.

\begin{figure}[!ht]
  \centering
  \includegraphics[width=\textwidth]{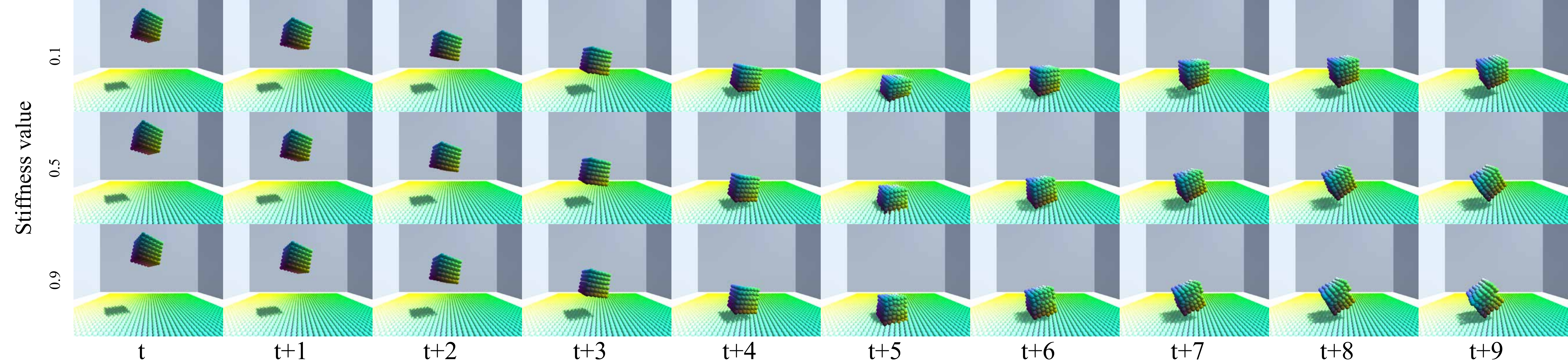}
   \caption{\textbf{Responsiveness to stiffness variations.} We vary the stiffness of a cube colliding with a planar surface. The top row shows a soft cube with stiffness value 0.1, the middle row a stiffness value of 0.5, and the bottom row a almost rigid cube with a stiffness value of 0.9. Our network responds as expected to the changing stiffness value deforming the soft cube stronger than the rigid cube.}
   \label{fig:var_stiff}
\end{figure}

In Figure \ref{fig:var_stiff} we demonstrate the model's ability to handle varying stiffness inputs. The network is trained on multiple soft bodies of varying stiffness. The stiffness values are obtained from FleX during dataset generation and vary between 0.1 and 0.9 for soft bodies. By manually changing the input stiffness during testing, we can produce predictions of objects with varying levels of rigidity. The decreasing level of deformation in frame $t+5$, from top to bottom, is consistent with the increasing stiffness.

\begin{figure}[!ht]
  \centering
  \includegraphics[width=\textwidth]{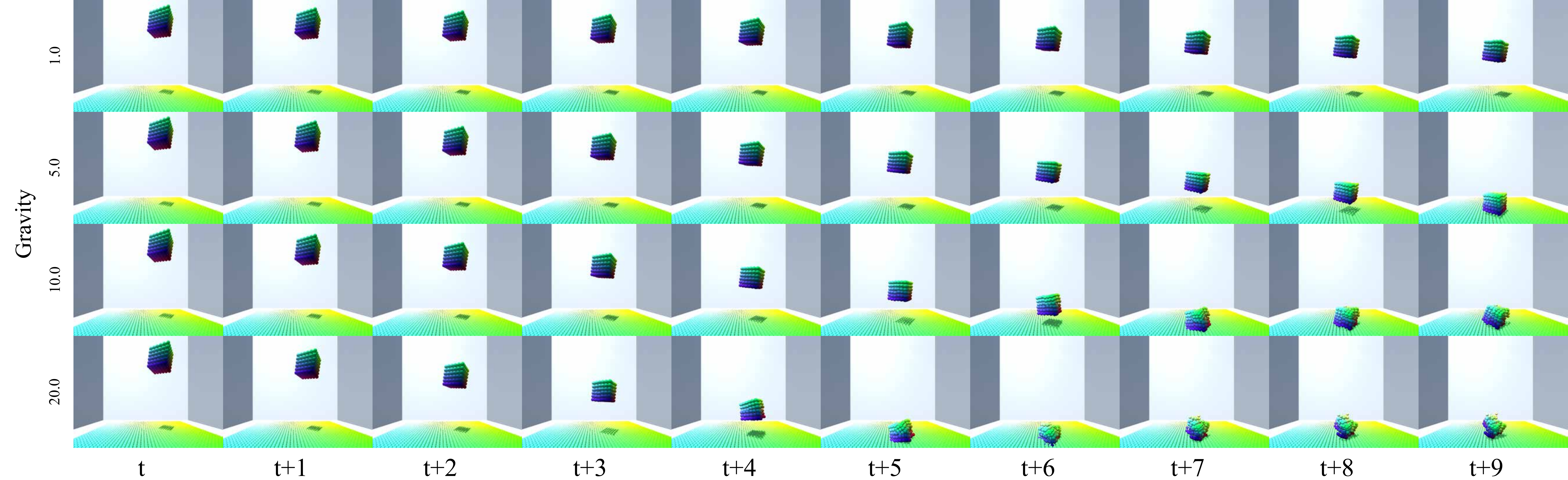}
   \caption{\textbf{Responsiveness to gravity variations.} We vary the gravity while a soft cube is falling on a planar surface. From top to bottom gravity values of 1.0, 5.0, 10.0 and 20.0 $m/s^2$ are depicted. We can see that the cube falls faster as gravity increases and even deforms the object when colliding with the floor under strong gravity. Our model behaves as expected when gravity changes.}
   \label{fig:var_grav}
\end{figure}

We also test whether the model can capture physical relationships in varying gravitational fields. Since the value of gravity is also an input to our model, we can train on data with a changing gravitational constant. Figure \ref{fig:var_grav} shows an example with four different gravitational constants, ranging from 1 to 20 $m/s^2$. As expected, the object falls faster with more gravity. 

\begin{figure}[!ht]
  \centering
  \includegraphics[width=\textwidth]{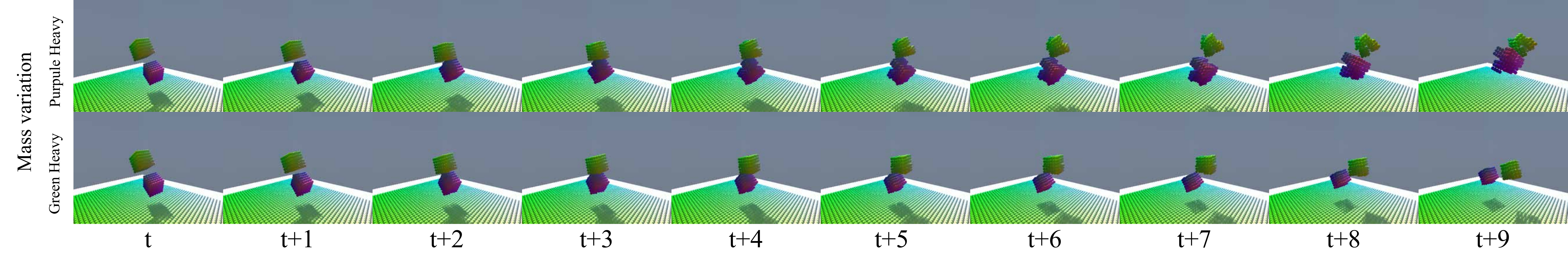}
   \caption{\textbf{Responsiveness to mass variations.} We vary the mass while two cubes collide. The top shows a scenario where the purple cube is heavy. Here the green cube bounces off stronger than the purple one. The bottom shows the same scenario but with green cube being heavy. Here the green cube doesn't move as strongly.}
   \label{fig:var_mass}
\end{figure}

As part of the particle state, we include the mass of each particle. While the total object mass is usually kept constant for most of the experiments, we test the case of varying mass by training on a dataset where the each object's mass will vary by a factor of up to three times. In Figure \ref{fig:var_mass} we manually increase the mass of one of the two objects in the collision and show that the heavier object is displaced less after the collision. 

\end{document}